\documentclass[lettersize,journal]{IEEEtran}
\usepackage[linesnumbered, ruled]{algorithm2e}
\usepackage{amsthm,amsmath,amssymb}
\usepackage{array}
\usepackage[caption=false,font=normalsize,labelfont=sf,textfont=sf]{subfig}
\usepackage{textcomp}
\usepackage{stfloats}
\usepackage{url}
\usepackage{verbatim}
\usepackage{graphicx}
\usepackage{booktabs,color}
\usepackage{cite}
\usepackage[affil-it]{authblk}
\usepackage{stackengine} 
\usepackage{mathtools}
\usepackage{mathrsfs}

\usepackage{indentfirst} 
\usepackage{multirow}
\usepackage[backref=page]{hyperref}
\usepackage{cleveref}
\usepackage{amssymb}
\usepackage{xcolor,eucal}
\usepackage{makecell}

\graphicspath{{figs}}
\let\oldnl\nl
\newcommand{\nonl}{\renewcommand{\nl}{\let\nl\oldnl}}
\newcommand{\hanxiR}{\mathbb{R}}
\newcommand{\II}{\text{I}}
\newcommand{\vv}[1]{\vec{\mathbf{#1}}}
\newcommand{\NL}{\\}

\DeclareMathOperator*{\argminB}{argmin}   

\SetKwRepeat{Do}{do}{while}%
\begin{document}

\title{Target before Shooting: Accurate Anomaly Detection and Localization under One
  Millisecond via Cascade Patch Retrieval}

\author[1,2,$\dagger$]{Hanxi Li\thanks{This work was mainly completed during Hanxi Li's visit to
    Zhejiang University.}}
\author[1,$\dagger$]{Jianfei Hu}
\author[3,$\star$]{Bo Li}
\author[2]{Hao Chen}
\author[4,$\star$]{Yongbin Zheng}
\author[2,$\star$]{Chunhua Shen}
\affil[1]{Jiangxi Normal University, Jiangxi, China}
\affil[2]{Zhejiang University, Zhejiang, China}
\affil[3]{Northwestern Polytechnical University, Shaanxi, China}
\affil[4]{National University of Defense Technology}
\affil[$\star$]{Corresponding author}
\affil[$\dagger$]{These authors contributed equally to this work}

\markboth{Journal of \LaTeX\ Class Files,~Vol.~14, No.~8, June~2023}%
{Shell \MakeLowercase{\textit{et al.}}: A Sample Article Using IEEEtran.cls for IEEE Journals}

\maketitle

\begin{abstract}

  %
  In this work, by re-examining the ``matching'' nature of Anomaly Detection (AD), we propose a
  new AD framework that simultaneously enjoys new records of AD accuracy and dramatically high running speed. In this framework, the anomaly detection problem is solved via a
  cascade patch retrieval procedure that retrieves the nearest neighbors for each test
  image patch in a coarse-to-fine fashion.
  %
  %
  Given a test sample, the top-$K$ most similar training images are first selected
  based on a robust histogram matching process. Secondly, the
  nearest neighbor of each test patch is retrieved over the similar geometrical locations on
  those ``global nearest neighbors'', by using a carefully trained local metric. Finally,
  the anomaly score of each test image patch is calculated based on the distance to its
  ``local nearest neighbor'' and the ``non-background'' probability. The proposed method is 
  termed ``Cascade Patch Retrieval'' (CPR) in this work. Different from the conventional
  patch-matching-based AD algorithms, CPR selects proper ``targets'' (reference images and locations)
  before ``shooting'' (patch-matching).
  %
  On the well-acknowledged MVTec AD, BTAD and MVTec-3D AD datasets, the proposed
  algorithm consistently outperforms all the comparing SOTA methods by remarkable margins, measured by various AD metrics.
  Furthermore, CPR is extremely efficient. It runs at the
  speed of $113$ FPS with the standard setting while its simplified version only
  requires less than $1$ ms to process an image at the cost of a trivial accuracy drop. The code of CPR is available at \href{https://github.com/flyinghu123/CPR}{https://github.com/flyinghu123/CPR}.

\end{abstract}


\section{Introduction}
\label{sec:intro}
To achieve industrial-level anomaly detection (AD) is challenging as the demanding
accuracy is high to ensure the reliability of defect inspection while the time budget is
limited on a running assemble-line.
Most of the recently proposed AD methods focus on increasing the recognition accuracy as it is difficult already, especially in
the standard setting where only normal samples are available for training.

The most straightforward way to realize the ``unsupervised'' AD is the ``one-class'' classification strategy
\cite{scholkopf2000support, scholkopf2001estimating}: by considering normal images or
patches as a single class and then the anomalous ones can be detected as outliers
\cite{ruff2018deep, yi2020patch, liznerskiexplainable2020, massoli2021mocca,
  defard2021padim, zhang2022pedenet, chen2022deep, dinh2014nice, rudolph2021same}.
Similarly, the Normalized-Flow-based approaches inherit the one-class assumption but
additionally impose a Gaussian distribution onto the class for better performance
\cite{yu2021fastflow, tailanian2022u, gudovskiy2022cflow, lei2023pyramidflow}.
In contrast to the single-class setting, some discriminative anomaly detectors
\cite{li2021cutpaste, zavrtanik2021draem, yang2023memseg} learn the pixel-wise binary
classifier with genuine normal samples and synthetic anomalous samples and can usually
lift the AD accuracy to some extent.
On the other hand, the AD methods based on distillation extract the ``knowledge'' of the
``teacher'' network, which is pre-trained on an AD-irrelevant dataset, to a ``student''
network, on only normal samples. The anomaly score of each image region is then
calculated based on the response difference between the two networks
\cite{salehi2021multiresolution, deng2022anomaly, bergmann2022beyond, zhang2023destseg}.
By considering the anomalous regions as ``occlusions'' or ''noises'', some researchers
propose to detect anomalies via image reconstruction and the anomaly scores are
positively related to the reconstruction residuals \cite{zong2018deep,
  dehaene2020anomaly, shi2021unsupervised, hou2021divide, wu2021learning}.
Compared with the sophisticated framework of other anomaly detectors, PatchCore
\cite{roth2022towards} offers a much simpler alternative for AD: it can achieve the
state-of-the-art AD accuracy via a plain patch matching/retrieval process. The success
of PatchCore inspires a number of its variants \cite{kim2022fapm, bae2022image,
  xie2023pushing, saiku2022enhancing, zhu2022anomaly} which try to increase the AD
performance mainly based on more informative patch features.
More recently, some AD methods propose to conduct an image alignment, explicitly or
implicitly, before detecting the anomaly \cite{huang2022registration, zheng2022focus,
  liu2023diversity}. In this way, the image patches can be only compared with the
(original or reconstructed) normal patches in a similar geometrical location. This
geometrical constraint usually leads to more reasonable matching distances and better AD
performances.

\begin{figure*}[t!]
  \begin{center}
    \includegraphics[width=0.95\textwidth]{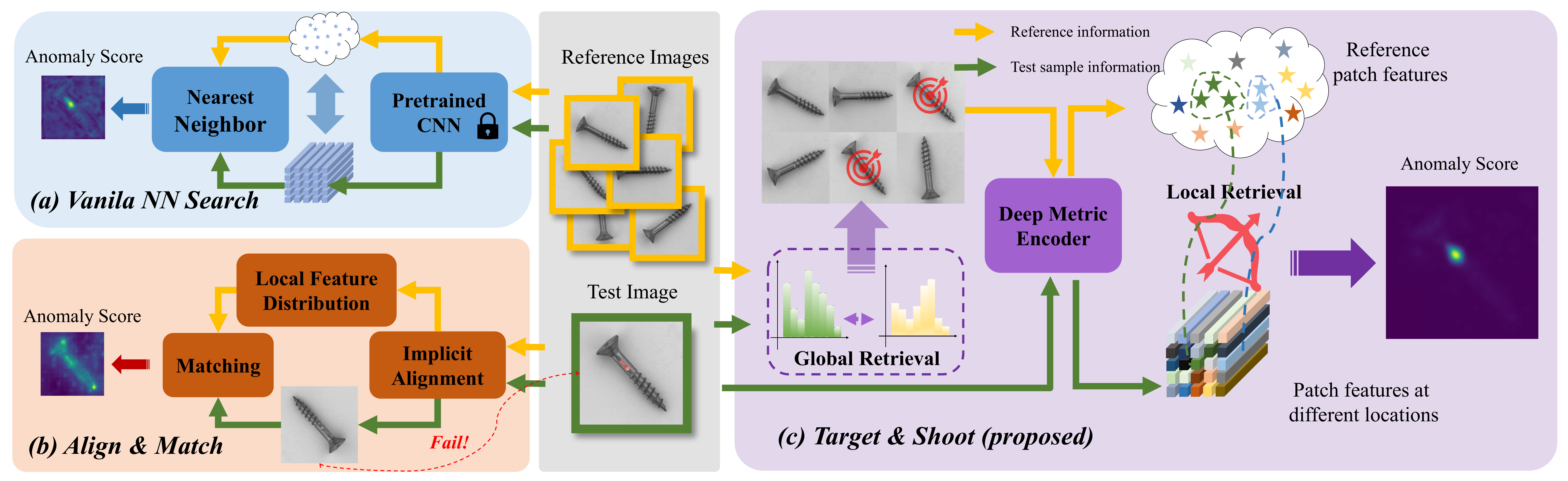}
    \caption{The main scheme of the proposed CPR algorithm. Left-top (a): the vanilla nearest
      neighbor searching method that searches the neighbors in a brute-force fashion;
      left-bottom (b): the ``align \& match'' strategy that perform an implicit alignment
      process before matching; right (c): the proposed ``target \& shoot'' scheme that realize
      AD within a retrieving cascade.}
    \label{fig:main_idea}
  \end{center}
\end{figure*}

In this work, by investigating the inherent ``matching'' character of AD, we propose a
more elegant framework to solve the AD problem. We found that a properly selected
reference set can significantly benefit the matching process, in terms of both
efficiency and accuracy. The main scheme of the proposed framework is shown in
\Cref{fig:main_idea}. As it is illustrated in the left part of the figure, the
vanilla nearest-neighbor searching methods retrieve the closest prototype for the test
patch in a brute-force way over the whole reference ``bank'' \cite{roth2022towards,
  kim2022fapm, bae2022image, xie2023pushing, saiku2022enhancing, zhu2022anomaly}. In this
way, an anomalous patch could be ``perfectly'' matched with a normal patch, which is
totally geometrically irrelevant to the query sample. Consequently, this unconvincing
patch matching/retrieval usually implies a failure of AD at the corresponding local
region.
The ``align \& match'' strategy \cite{huang2022registration, zheng2022focus,
  liu2023diversity} seems a better alternative as the images are firstly aligned to
generate geometrically meaningful matching pairs. However, when the alignment fails on
the test image, the geometrical constraint could cause even worse AD results.
On the contrary, we perform the retrieval in a ``target \& shoot'' fashion. As we can
see in the right part of \Cref{fig:main_idea}, instead of aligning the test image, we
select proper reference ``targets'' for each test patch. In specific, the qualified target samples should locate at similar image coordinates to the test patch and should be only extracted from the
reference images obtained in the global retrieval stage. The local retrieval is conducted in the feature space which is learned
using a carefully customized contrastive loss and enjoys high accuracy. In addition,
differing from the ``align \& match'' strategy, the cascade scheme is naturally
robust to global retrieval errors, and thus the high recall rate of retrieval is ensured.

We term the proposed method
``Cascade Patch Retrieval'' (CPR) because the retrieval is
realized with a two-stage cascade process. For each test patch, our CPR model first
selects qualified reference samples in the oracle and then matches the query patch with
the selected references. In metaphorical words, the CPR ``targets'' on the proper
references before ``shooting''. In the extensive experiments of this work, the proposed
``target \& shoot'' methodology illustrates remarkably high AD accuracy as well as time
efficiency. In summary, the contribution of this work is threefold, as listed below.
\begin{itemize}
  \item
        Firstly, we cast the AD task as a cascade retrieval problem. Instead of brute force
        searching over the patch bank or aligning the test image to a ``standard'' pose, the
        cascade retrieval strategy naturally possesses a high retrieving recall rate and
        offers sufficient room for learning better retrieval metrics.
  \item
        Secondly, to address the over-fitting problem under the few sample condition for
        most AD datasets, we propose a novel metric learning framework with a customized
        contrastive loss and a more conservative learning strategy. The yielded CPR model
        outperforms existing SOTA algorithms by large margins, in three
        well-acknowledged AD datasets, namely MVTec AD, MVTec-3D AD and BTAD.
  \item
        Finally, in this work, the original CPR algorithm is smartly simplified for higher
        efficiency. The fastest versions of CPR, \emph{i.e.}, CPR-Faster runs at a speed
        over $1000$ FPS while still maintaining accuracy superiority over most current SOTA
        methods.
\end{itemize}

The rest of this paper is organized as follows: in \Cref{sec:related}, the related work
of the proposed algorithm is introduced; the details of the CPR algorithm are represented
in \Cref{sec:method} and we report all the experiment results in \Cref{sec:experiments};
the \Cref{sec:conclusion} summarizes this paper and discusses the future work of CPR.

\section{Related work}
\label{sec:related}

\subsection{Anomaly Detection via Patch Matching}
\label{subsec:matching}

A simple way to detect and localize anomalies is to find a normal prototype for each test
patch and assign high anomaly scores to those with low similarities. As a typical and
simple example, the PatchCore algorithm \cite{roth2022towards} proposes to build a
``memory bank'' of patch features based on the coreset-subsampling algorithm and the anomaly
score is then estimated according to the matching distance between the test patch
and its nearest neighbor. Interestingly, merely with this simple matching strategy, PatchCore
achieves dramatically high performance on the well-acknowledged MVTech-AD dataset
\cite{bergmann2019mvtec}.
The success of PatchCore inspired a number of following works: the FAPM algorithm
\cite{kim2022fapm} designs patch-wise adaptive coreset sampling to improve the matching
speed.  \cite{bae2022image} improves the matching quality by employing the position and
neighborhood information of the test patch. Graphcore \cite{xie2023pushing} proposes a
graph representation to adapt PatchCore to the few-shot scenario.
Apart from the vanilla brute-force matching, some recently proposed methods
\cite{huang2022registration, zheng2022focus, liu2023diversity} try to conduct the patch
matching between the test patch and the reference set with similar geometrical
properties. In a straightforward way, the test image is firstly aligned to a
``standard'' pose using rigid \cite{huang2022registration, zheng2022focus} or nonrigid
\cite{liu2023diversity} geometrical deformations which could be explicitly realized via
one or more learnable neural network layers. The aligned deep feature tensor are matched
with a reconstructed prototype \cite{liu2023diversity} or a position-conditioned
distribution \cite{huang2022registration, zheng2022focus} and the anomaly score is
estimated according to the deformation magnitude from the prototype
\cite{liu2023diversity} or the Mahalanobis distances from each test patch to the
corresponding distribution \cite{huang2022registration, zheng2022focus}.

Despite the well-designed framework, the ``align \& matching'' strategy does not
illustrate significant improvement over the non-aligned patch matching methods on the
well-acknowledged AD datasets, partially due to the failure cases of image alignment. In
this work, we impose the geometrical constraints in another way. Instead of aligning
the image, we directly match each test patch with its nearest neighbor, given certain
constraints. The constrained searching process is realized in a two-stage
retrieval cascade and new SOTA AD performance is obtained.

\subsection{Deep Learning based Image Retrieval}
\label{subsec:retrieval}

Image retrieval is a long-standing and fundamental task in computer vision and
multimedia.  Typically, one needs to find the nearest neighbor(s) of a ``query'' image
over the image ``gallery'', which is an image oracle for matching in fact. In this way,
the unknown property (\emph{e.g.} category label or instance identity) of the query
image can be determined by the matched nearest neighbors \cite{chen-survey-2022deep}. A great
amount of effort has been devoted to realizing better image retrieval algorithms. Most
existing image retrieval methods focus on obtaining semantic-aware global features which
could be aggregated from the off-the-shelf models \cite{2015Cross, 2015Exploiting,
  2017Unsupervised} or specifically learned models \cite{2015NetVLAD, 2016Deep,
  2021Instance}. In particular, to ensure a swift image retrieval process on large-scale
datasets, many approaches learn binary codes, instead of conventional real-valued
features, to realize the hashing retrieval \cite{2016Accurate, Mor2016Nested, 2017Deep,
  2018Supervised}, which is highly efficient in terms of both memory usage and time
consumption.

A frequently employed strategy of state-of-the-art retrieval algorithms is the
``re-ranking'' scheme that consists two phases, \emph{i.e.}, the initial ranking stage
and re-ranking stage respectively \cite{2016Image, 2018Detect, 2021Instance}. In
specific, the top-k nearest neighbors are first selected for the query image from a
global view in the former stage and the retrieval order is then modified according to
the matching state between the local descriptors extracted from the query image and its
neighbors. In this work, inspired by the high-level concept of this sophisticated
strategy, we smartly transfer the conventional matching operation of anomaly detection
into a cascade retrieval problem which is also solved in a coarse to fine manner.

\section{The proposed method}
\label{sec:method}

\subsection{The overview of the proposed method}
\label{subsec:alg_overview}
\begin{figure*}[htbp]
  \centering
  \includegraphics[width=6.8in]{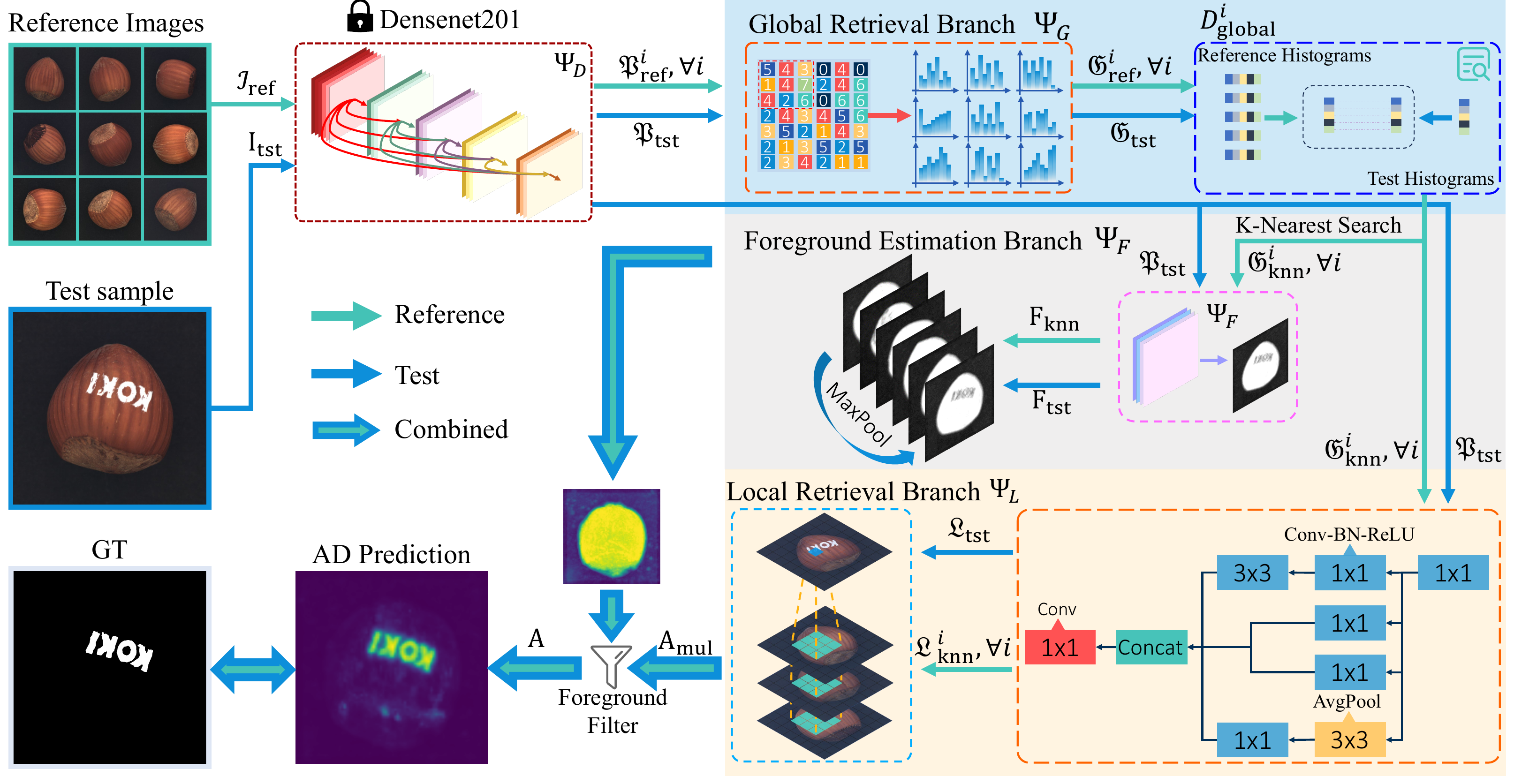}
  \caption{CPR train procedure for detecting and localizing anomalies on
    images. The whole process could be divided into $4$ sectors, namely the DenseNet$201$
    backbone, the global retrieval branch, the local retrieval branch and the foreground
    estimation branch, respectively.}\label{fig:our_method}
\end{figure*}

The detailed inference scheme of the proposed Cascade Patch Retrieval (CPR) algorithm is
shown in \Cref{fig:our_method}.
According to the illustration, the input of the CPR algorithm includes a test image
$\II_{\text{tst}} \in \hanxiR^{H_I\times W_I\times 3}$ and a reference image set
$\mathcal{I}_{\text{ref}} \triangleq \{\II^i_{\text{ref}} \in \hanxiR^{H_I\times W_I\times
    3} \mid i = 1, 2, \dots, N_R\}$. The CNN-based CPR model consists of $4$ subnetworks, namely
the DenseNet$201$ \cite{huang2018densely} backbone, the Global Retrieval Branch (GRB), the
Local Retrieval Branch (LRB) and the Foreground Estimation Branch (FEB), respectively. As
the raw feature generator, the DenseNet$201$ can be denoted as the following function
\begin{equation}
  \label{equ:raw_fea}
  \Psi_{D}: \II  \in \hanxiR^{H_I\times W_I\times 3} \rightarrow \mathfrak{P}
  \in \hanxiR^{H_P \times W_P \times C_P},
\end{equation}
where $\mathfrak{P}$ denotes the generated raw feature which will be fed into the
succeeding three branches. In a similar manner, the global retrieval branch
$\Psi_{G}(\cdot)$, the local retrieval branch $\Psi_{L}(\cdot)$ and the foreground
estimation branch $\Psi_{F}(\cdot)$ are defined as follows
\begin{equation}
  \label{equ:branch_functions}
  \begin{split}
    & \Psi_{G}: \mathfrak{P} \in \hanxiR^{H_P \times W_P \times C_P} \rightarrow
    \mathfrak{G} \in \hanxiR^{S \times S \times N_C}\\
    & \Psi_{L}: \mathfrak{P} \in \hanxiR^{H_P \times W_P \times C_P} \rightarrow
    \mathfrak{L} \in \hanxiR^{H_L \times W_L \times C_L} \\
    & \Psi_{F}: \mathfrak{P} \in \hanxiR^{H_P \times W_P \times C_P} \rightarrow \text{F} \in
    \hanxiR^{H_F \times W_F}
  \end{split}
\end{equation}
note that the outputs of $\Psi_{G}(\cdot)$ is a collection of $S^2$ histograms with $N_C$
bins, $\Psi_{L}(\cdot)$ generates a $3$-D tensor while the output of $\Psi_{F}(\cdot)$ is
a $2$-D confidence map $\text{F}$.

The first stage of the cascade retrieval is conducted based on the ``global feature''
$\mathfrak{G}$ and the top-$k$ nearest neighbors of the test image $\II_{\text{tst}}$ is
stored into the image set $\mathcal{N}_{\text{ref}}$.

The second stage of the cascade retrieval is performed based on the local feature
$\mathfrak{L} \in \hanxiR^{H_P \times W_P \times C_P}$ which can be viewed as a collection
of $H_P\cdot W_P$ vector features with each element corresponding to an image patch.
The primitive anomaly score of each test patch is estimated according to the patch feature distance between itself and its nearest neighbor retrieved from $\mathcal{N}_{\text{ref}}$ and at similar image locations.

Finally, under the assumption that the anomalies only take place in the foreground part,
the primitive anomaly scores are corrected via using the foreground confidence map
$\text{F} \in \hanxiR^{H_F \times W_F}$ predicted by the foreground estimation branch
$\Psi_F(\cdot)$.

\subsection{The global retrieval branch}
\label{subsec:grb}
The Global Retrieval Branch (GRB) is actually a statistical feature generator. Given $N_R$
reference (anomaly-free) images $\{\II^i_{\text{ref}} \in \hanxiR^{H_I\times W_I\times 3}
  \mid i = 1, 2, \dots, N_R\}$, and the corresponding DenseNet feature tensors
$\{\mathfrak{P}^i_{\text{ref}} = \Psi_D(\II^i_{\text{ref}}) \mid i = 1, 2, \dots, N_R\}$, we
firstly flatten each tensor to build the ``raw patch feature'' set as
\begin{equation}
  \label{equ:raw_patch_feature}
  \mathfrak{P}^i_{\text{ref}} \xrightarrow{\texttt{flatten}} \{\vv{p}^{i, j}_{\text{ref}}
  \in \hanxiR^{C_P} \mid j = 1, 2, \dots, H_P W_P\}
\end{equation}
Then suppose all the raw patch features belonging to different reference images are
collected together into the feature set $\mathcal{P}_{\text{ref}} \triangleq
  \{\vv{p}^{u}_{\text{ref}} \in \hanxiR^{C_P} \mid u = 1, 2, \dots, N_R H_P W_P\}$, a
$K$-means clustering method is performed on $\mathcal{P}_{\text{ref}}$ to obtain $N_C$
clustering centers $\mathcal{C}_{\text{ref}} \triangleq \{\vv{c}^v_{\text{ref}} \in
  \hanxiR^{C_P} \mid v = 1, 2, \dots, N_C\}$.

The block-wise statistics of the test image $\II_{\text{tst}}$ can be obtained based
on the raw feature tensor $\mathfrak{P}_{\text{tst}}$ and the cluster center set
$\mathcal{C}_{\text{ref}}$. In specific, the tensor $\mathfrak{P}_{\text{tst}}$ is evenly
divided into $S \times S$ sub-tensors as
\[
  \hat{\mathfrak{P}}^{u, v}_{\text{tst}} \in
  \hanxiR^{\frac{H_P}{S}\times\frac{W_P}{S}\times C_P}, u,v = 1, 2, \dots, S.
\]
For each sub-tensor, we extract its statistics in the ``Bag of Words'' (BoW) style as
\begin{equation}
  \label{equ:statistic}
  \begin{split}
    \hat{\mathfrak{P}}^{u, v}_{\text{tst}} & \xrightarrow{\texttt{flatten}}
    \mathcal{P}_{\text{tst}} \triangleq \{\vv{p}^{i}_{\text{tst}} \mid i = 1,
    \dots, \frac{H_P W_P}{S^2}\} \\
    & \xrightarrow{\texttt{BoW}} \vv{h}_{u, v} \in \hanxiR^{N_C}, \forall u, v
  \end{split}
\end{equation}
where $\vv{h}_{u, v}$ denotes a BoW histogram of $\hat{\mathfrak{P}}^{u, v}_{\text{tst}}$
with respect to the codebook $\mathcal{C}_{\text{ref}}$ and this histogram is normalized
so that $\|\vv{h}_{u, v}\|_{l_1} = 1, \forall u, v$. The histograms
of a reference image $\II^i_{\text{ref}}$ can also be obtained in a similar manner as
\Cref{equ:statistic} and let us denote them as $\vv{h}^{\text{ref}}_{u, v, i}$. Let
$D^{u,v,i}_{\text{KL}} \triangleq \text{KL}(\vv{h}^{\text{ref}}_{u, v, i}, \vv{h}_{u, v})$
denote the Kullback-Leibler (KL) divergence between two block histograms with the same
block ID, one can get the following sorted (in increasing order) block divergences
\[
  \{D^i_1 \leqslant D^i_2 \leqslant D^i_3 \dots \leqslant D^i_{S^2}\}.
\]
Then the global features distance between $\II_{\text{tst}}$ and $\II^i_{\text{ref}}$ is
estimated as
\begin{equation}
  \label{equ:global_dist}
  D^i_{\text{global}} \triangleq \frac{1}{S^2 - \tau}\sum_{j = 1}^{S^2 - \tau} D^i_j,
  ~\forall i = 1, 2, \dots, N_R,
\end{equation}
where $\tau$ is a small number for ignoring large block-wise KL divergences so that the
overall distance estimation is robust to partial image contamination.

Based on the global feature distance, the top-$K$ neighbor reference images are retrieved
for $\II_{\text{tst}}$ and stored into the image set $\mathcal{N}_{\text{ref}} \triangleq
  \{\II^1_{\text{knn}}, \II^2_{\text{knn}}, \dots, \II^K_{\text{knn}}\}$ which are employed
as the reference images in the following patch retrieval process.

In practice, this searching process over the reference images usually lead to a set of
``pseudo-aligned'' nearest neighbors (see \Cref{subsec:qualitative} for details). Thus our
global retrieval branch offers an alternative to the image alignment approaches
\cite{huang2022registration, zheng2022focus, liu2023diversity}, while with higher
simplicity and accuracy.

\subsection{The local retrieval branch}
The Local Retrieval Branch (LRB) plays the most important role in the CPR algorithm.
\subsubsection{The structure of LRB}
As to the branch structure, inspired by the pioneering work \cite{szegedy2015rethinking},
we modify the well-known inception block to build our LRB. As it is shown in
\Cref{fig:our_method}, the input feature (\emph{i.e.}, the raw feature tensor
$\mathfrak{P}$ of \Cref{equ:raw_fea}) are processed by the block via four main paths, with
different combinations of the ``Conv-BN-ReLU'' blocks and the average pooling layers.  The
output tensors of these four paths are firstly concatenated together and then further
processed by a $1 \times 1$ convolutional layer. According to \Cref{equ:branch_functions},
the final output tensor of the test image $\II_{\text{tst}}$ is given by
\begin{equation}
  \label{equ:image_to_L}
  \mathfrak{L}_{\text{tst}} = \Psi_L(\Psi_D(\II_{\text{tst}})) \in \hanxiR^{H_L \times W_L
    \times C_L}.
\end{equation}

\subsubsection{The inference process of LRB}
Let us flatten the tensor $\mathfrak{L}_{\text{tst}}$ into a set of feature vectors with
the associated row-column coordinates as
\begin{equation}
  \label{equ:local_patch_feature}
  \mathfrak{L}_{\text{tst}} \xrightarrow{\texttt{flatten}} \mathcal{L}_{\text{tst}}
  \triangleq \{\vv{l}^{r, c}_{\text{tst}} \in \hanxiR^{C_L} \mid \forall r, c\},
\end{equation}
where $r = 1, 2, \dots, H_L$, $c = 1, 2, \dots, W_L$.

Given the nearest neighbor set $\mathcal{N}_{\text{ref}}$ obtained by GRB, one can calculate the
corresponding local feature tensors as $\{\mathfrak{L}^j_{\text{knn}} =
  \Psi_L(\Psi_D(\II^j_{\text{knn}})) \in \hanxiR^{H_L \times W_L \times C_L} \mid j = 1, 2, \dots,
  K\}$. In a similar way to \Cref{equ:local_patch_feature}, the feature vectors of
$\mathcal{N}_{\text{ref}}$ writes
\begin{equation}
  \label{equ:local_patch_feature_ref}
  \{\mathfrak{L}^j_{\text{knn}} \mid \forall j\} \xrightarrow{\texttt{flatten}}
  \mathcal{L}_{\text{knn}} \triangleq \{\vv{l}^{r, c, j}_{\text{knn}} \in \hanxiR^{C_L}
  \mid \forall r, c, j\},
\end{equation}
The local patch retrieval is then conducted between each feature vector in
$\mathcal{L}_{\text{tst}}$ and the patch reference set $\mathcal{L}_{\text{knn}}$. In
particular, given a test patch feature $\vv{l}^{r, c}_{\text{tst}}$, its nearest neighbor
is retrieved as
\begin{equation}
  \label{equ:loal_retrieval}
  \vv{l}^{\ast}_{r, c} = \argminB_{\substack{\forall |\bar{r} - r| < \delta \\ \forall
      |\bar{c} - c| < \delta \\ \forall j = 1, \dots, K}}
  \left[1 - \Phi_{\text{cos}}(\vv{l}^{r, c}_{\text{tst}}, \vv{l}^{\bar{r}, \bar{c},
      j}_{\text{knn}})\right],
\end{equation}
where $\Phi_{\text{cos}}(\cdot)$ stands for the function of cosine similarity between two
vectors. Accordingly, the minimal distance is defined as:
\begin{equation}
  \label{equ:min_dist}
  \text{A}_{\text{ori}}[r, c] = d^{\ast}_{r, c} \triangleq 1 -
  \Phi_{\text{cos}}(\vv{l}^{r, c}_{\text{tst}}, \vv{l}^{\ast}_{r,
    c}),
\end{equation}
where $\text{A}_{\text{ori}} \in \hanxiR^{H_L \times W_L}$ denotes the map of anomaly
score predicted by the LRB.

\subsubsection{The training strategy of LRB}
\label{subsubsec:lrb_train}
To achieve a good metric function $\Psi_{\text{L}}$ for patch retrieval, we learn the
parameters of LRB using a sophisticated training scheme and a modified contrastive loss
\cite{1640964}.

In particular, firstly, a query image $\II_{\text{que}}$ and a reference
image $\II_{\text{ref}}$ randomly selected from the $K$-NN set $\mathcal{N}_{\text{ref}}$
are employed as the input of the metric learning process. Following the data
augmentation strategy of \cite{yang2023memseg}, pseudo anomalies are randomly
merged with $\II_{\text{que}}$ to generate the synthetically defective image
$\tilde{\II}_{\text{que}}$.
Secondly, $\tilde{\II}_{\text{que}}$ and $\II_{\text{ref}}$ are processed by
\Cref{equ:image_to_L} to obtain the feature tensors $\mathfrak{L}_{\text{que}}$ and
$\mathfrak{L}_{\text{ref}}$ which are further flattened to the vector feature sets
$\mathcal{L}_{\text{que}} \triangleq \{\vv{l}^{r, c}_{\text{que}} \in \hanxiR^{C_L} \mid
  \forall r, c\}$ and $\mathcal{L}_{\text{ref}} \triangleq \{\vv{l}^{r, c}_{\text{ref}} \in
  \hanxiR^{C_L} \mid \forall r, c\}$ respectively.

Thirdly, the training samples of the metric learning, \emph{i.e.}, the feature pairs are
sampled using Algorithm~\ref{alg:sample}. Note that three kinds of feature pairs are
sampled, namely the positive pairs, the remote pairs and the anomalous pairs. The former
one is positive ($y_i = 1$) and the latter two kinds are both negative ($y = 0$) but
differ in the reason for being negative.


\begin{algorithm}[ht]
  \caption{Feature-pair sampling strategy of CPR}  
    \label{alg:sample}
    \LinesNumbered  
    \KwIn{
      Query feature set $\mathcal{L}_{\text{que}}$; reference feature set
      $\mathcal{L}_{\text{ref}}$; binary mask $\text{M}_{\text{que}} \in \mathbb{B}^{H_L
      \times W_L}$ of the synthetic anomaly of $\tilde{\II}_{\text{que}}$; number of
      samples $\Gamma$; distance threshold $\theta$; vector normalization function
      $\Phi(\cdot)$.
    }  
    \KwOut{
    The set of query-reference feature pairs $\Xi = \{\xi_i = [\vec{\boldsymbol{\eta}}^i_{\text{que}},
      \vec{\boldsymbol{\eta}}^i_{\text{ref}}] \mid i = 1, \dots, \Gamma\}$; Binary labels $\vv{y} \in
      \mathbb{B}^{\Gamma}$.
    }
    \textbf{Algorithm:}\\
    $\Xi \leftarrow \{\}$, $\vv{y} = \mathbf{0}, i = 1$\\
    \While{$i \leqslant \Gamma$}
    {
        \tcc{{positive pairs sampling}}
        $[r_{q},~c_{q}] = \texttt{RandSelect}([\![ \text{M}_{\text{que}} = = 0]\!])$\\
        $[r_{r},~c_{r}] = [r_{q},~c_{q}]$\\
        $\Xi \leftarrow \xi_i = [\Phi(\vv{l}^{r_q, c_q}_{\text{que}}), \Phi(\vv{l}^{r_r,
        c_r}_{\text{ref}})]$, $\vv{y}[i] = 1$ \\

        \tcc{{remote pairs sampling}}
        $[r_{q},c_q] = \texttt{RandSelect}([\![\|r_r - r_q, c_r - c_q\|_{l_2}> \theta]\!])$\\
          $\Xi \leftarrow \xi_i = [\Phi(\vv{l}^{r_q, c_q}_{\text{que}}), \Phi(\vv{l}^{r_r,
          c_r}_{\text{ref}})]$, $\vv{y}[i + 1] = 0$ \\
%

        \tcc{{anomalous pairs sampling}}
        $[r_{q},~c_{q}] = \texttt{RandSelect}([\![ \text{M}_{\text{que}} = = 1]\!])$\\
        $[r_{r},~c_{r}] = [r_{q},~c_{q}]$\\
        $\Xi \leftarrow \xi_i = [\Phi(\vv{l}^{r_q, c_q}_{\text{que}}), \Phi(\vv{l}^{r_r,
        c_r}_{\text{ref}})]$, $\vv{y}[i + 2] = 0$ \\

        $i = i + 3$ \\
    }
\end{algorithm}

Fourthly, in Algorithm~\ref{alg:sample}, a remote pair is considered as negative only
because of the coordinate distance between the two involved features. In this way,
unnecessary ambiguities could be brought into the training process due to the existence of
the ``remote but similar'' feature pairs.
In this work, we propose to reduce the training sample weights of those ambiguous pairs.
Concretely, given the row-column coordinates of the two involved features in a pair
$\xi_i$ writes $[r_i, c_i, \hat{r}_i, \hat{c}_i]$, $\omega_i$ and the raw feature tensors
of query and reference images are $\mathfrak{P}_{\text{que}} =
  \Psi_D(\tilde{\II}_{\text{que}})$ and $\mathfrak{P}_{\text{ref}} =
  \Psi_D(\II_{\text{ref}})$ \footnote{Note that here the tensors $\mathfrak{P}_{\text{que}}$
and $\mathfrak{P}_{\text{ref}}$ are resized so that their height and width are identical to
$\mathfrak{L}_{\text{que}}$ and $\mathfrak{L}_{\text{ref}}$}. The weight of $\xi_i$ is
denoted as $\omega_i$ which is calculated as
\begin{equation}
  \label{equ:sample_w}
  \omega_i =
  \begin{cases}
    \|\vv{p}^{r,c}_{\text{que}} - \vv{p}^{\hat{r},\hat{c}}_{\text{ref}}\|_{l_2} / \delta &
    \text{if }\xi_i~\text{is a remote pair}                                                                 \\
    1                                                                                    & \text{otherwise}
  \end{cases}
\end{equation}
where $\vv{p}^{r,c}_{\text{que}}$ and $\vv{p}^{\hat{r},\hat{c}}_{\text{ref}}$ are the
sliced feature vectors from $\mathfrak{P}_{\text{que}}$ and $\mathfrak{P}_{\text{ref}}$,
according to the coordinates $[r_i, c_i, \hat{r}_i, \hat{c}_i]$,  $\|\cdot\|_{l_2}$ stands
for the $l_2$ norm function and $\delta$ is the pre-estimated average raw feature distance
between randomly selected reference and query images.

Finally, the modified contrastive loss is defined as
\begin{equation}
  \label{equ:contrastive}
  \begin{split}
    \mathsf{L}_{\text{contrast}} = & \frac{1}{\Gamma}\sum^{\Gamma}_{i = 1}\omega_i \cdot
    [y_i\cdot \text{max}(0, m^{+} - {\vec{\boldsymbol{\eta}}^{i~\text{T}}_{\text{que}}}
      \vec{\boldsymbol{\eta}}^i_{\text{ref}}) + \\
      & (1 - y_i)\cdot \text{max}(0, \vec{\boldsymbol{\eta}}^{i~\text{T}}_{\text{que}}
      \vec{\boldsymbol{\eta}}^i_{\text{ref}} - m^{-})]^p,
  \end{split}
\end{equation}
where the inner product between $\vec{\boldsymbol{\eta}}^i_{\text{que}}$ and
$\vec{\boldsymbol{\eta}}^i_{\text{ref}}$ reflects the pair similarity, $m^{+}$ and
$m^{-}$ are the hyper-parameters of ``similarity threshold'' for positive pairs and
negative pairs, $p$ is a constant set larger than $1$ to emphasis the hard pair samples in
training.

\begin{figure*}[htbp]
  \centering
  \includegraphics[width=7in]{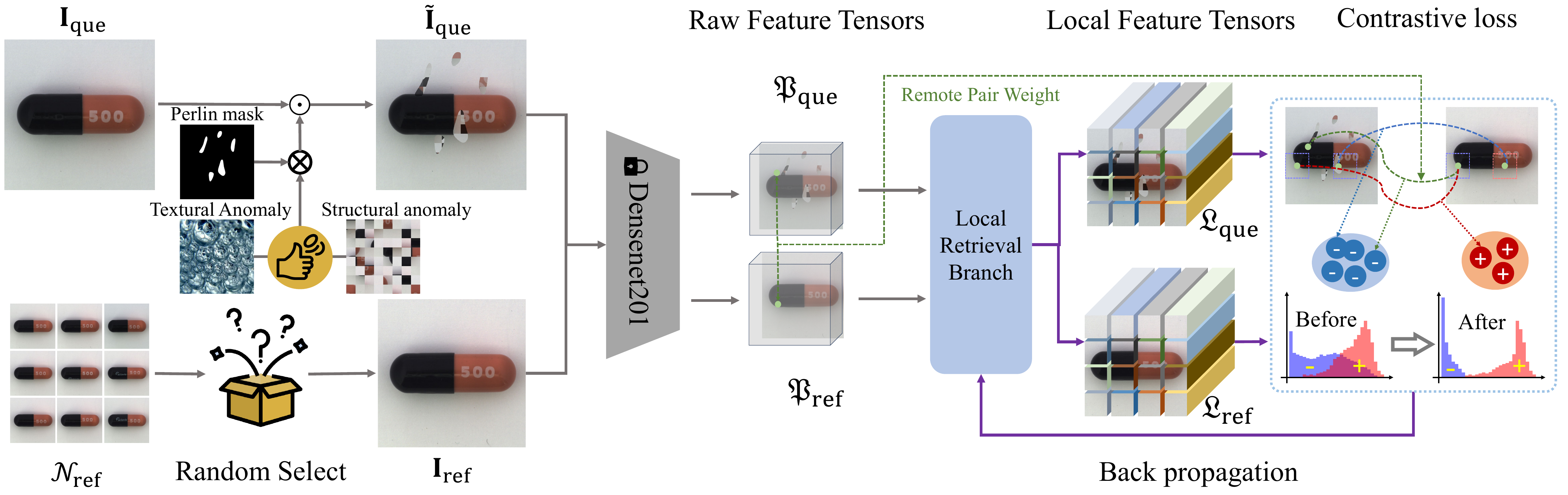}
  \caption{
    CPR inference procedure for detecting and localizing anomalies on images. Note that
    the remote feature pairs (green) and the anomalous feature pairs (blue) are labeled as
    negative while the positive feature pairs (red) are extracted from the identical and
    defect-free location on two feature tensors.}
  \label{fig:train}
\end{figure*}

\subsection{The foreground estimation branch}
\label{subsec:foreground}
In many real-life scenarios, the inspected object (\emph{e.g.} a screw or a hazelnut
\cite{bergmann2019mvtec}) is photographed with a large surrounding background region and
without explicit annotations. Recalling that the concerned anomalies are all located on the
object part, removing the ``false alarms'' on the background region can therefore
effectively increase the AD accuracy. Consequently, in some recently proposed works
\cite{yang2023memseg, zhang2022prototypical, wang2023multimodal, yao2022explicit},
foreground/background information is well exploited for anomaly detection.

\begin{figure}[htbp]
  \centering
  \includegraphics[width=3.2in]{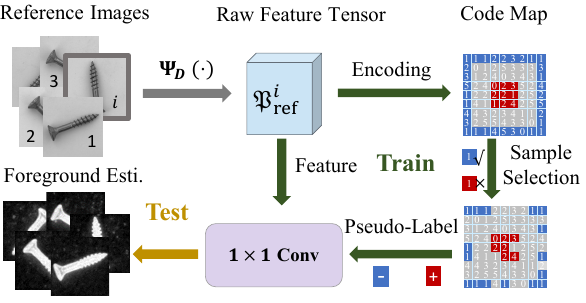}
  \caption{The foreground estimation strategy of the proposed CPR method. The deep
    features obtained by DenseNet$201$ and the pseudo-labels are used to learn a $1 \times
      1$ convolutional layer.}
  \label{fig:foreground}
\end{figure}

In this paper, we employ a Foreground Estimating Branch (FEB) to classify
the pixels as foreground or background. The workflow of FEB is depicted in
\Cref{fig:foreground}. As it is shown in the figure, firstly, each training image
$\II^i_{\text{ref}}$ is processed by DenseNet$201$ backbone as \Cref{equ:raw_fea} to
generate the raw feature tensor $\mathfrak{P}^i_{\text{ref}} \in \hanxiR^{H_P \times W_P
    \times C_P}$. Secondly, all the vector feature $\vv{p}^{i,j}_{\text{ref}} \in
  \hanxiR^{C_P}, \forall j$ of this tensor are coded with the codebook
$\mathcal{C}_{\text{ref}}$ defined in \Cref{subsec:grb} and then the codes are recombined
into a code map $\text{C}^i_{\text{ref}} \in \hanxiR^{H_P \times W_P}$ (shown as numbers on the right column of \Cref{fig:foreground}), which is mainly used to select training samples for FEB.

In this work, some of the vector features $\vv{p}^{i,j}_{\text{ref}} \in \hanxiR^{C_P},
  \forall j$ of this tensor are selected as training features of the classifier. As there is
no explicit foreground/background label in most AD datasets \cite{bergmann2019mvtec,
  Bergmann_2022}, we propose to generate the pseudo-labels for the selected deep features.
In general, the surrounding area (shown in blue in the top-right block of
\Cref{fig:foreground}) are treated as negative (background) regions and the center area
(shown in red in the two blocks) are the positive (foreground) region. However, to further
remove the potential class ambiguity, only the surrounding features with the majority code
(cluster-$1$ in the example of \Cref{fig:foreground}) are selected as negative samples and
the center samples with this majority code are abandoned. The training samples of the
FEB is selected over all the reference images and a $1 \times 1$
convolutional layer is employed to classify the vector features
$\vv{p}^{i,j}_{\text{ref}}, \forall j$ into the two categories, as shown in the
bottom-left corner of \Cref{fig:foreground}.

In practice, suppose that the foreground prediction map of the test image $\II_{\text
    {tst}}$ is denoted as $\text{F}_{\text{tst}} \in \hanxiR^{H_F \times W_F}$ and those maps
of $\II_{\text {tst}}$'s $K$ nearest neighbors are collected in the set $\mathcal{F}
  \triangleq \{\text{F}^i_{\text{ref}}  \in \hanxiR^{H_F \times W_F} \mid i = 1, 2, \dots,
  K\}$, each element of the final foreground estimation map $\text{F}^{\ast} \in
  \hanxiR^{H_F \times W_F}$ is given by
\begin{equation}
  \label{equ:map_tensor}
  \text{F}^{\ast}[r, c] = \max\left(\text{F}_{\text{tst}}[r, c], ~\max_{i =
    1, \dots, K}  \text{F}^i_{\text{ref}}[r, c]\right),
\end{equation}
where $r = 1, \dots H_F$ and $c = 1, \dots, W_F$.
The final anomaly prediction map $\text{A}^{\ast} \in \hanxiR^{H_F \times W_F}$ is then
obtained as
\begin{equation}
  \label{equ:final_anomaly_map}
  \text{A}^{\ast} = \texttt{UpInterp}(\text{A}_{\text{ori}}) \odot \text{F}^{\ast},
\end{equation}
where $\odot$ denotes the element-wise multiplication and function
$\texttt{UpInterp}(\cdot)$ stands for the up-sampling operation as $\text{A}_{\text{ori}}
  \in \hanxiR^{H_L \times W_L}$ and usually $H_F \geqslant H_L ~\& ~ W_F \geqslant W_L$.
Note that the proposed FEB is not compatible to all the AD datasets, \emph{e.g.} the
texture sub-categories of MVTec AD \cite{Bergmann_2019_CVPR}. In these scenarios, we just
disable the FEB for the whole learning-inference process.

\subsection{The end-to-end inference process of CPR}
\label{subsec:cpr_learn}
The end-to-end inference process of the proposed method is shown in
Algorithm~\ref{alg:infer}. Note that in practice the proposed method is designed in a
multi-scale style. In particular, two raw feature tensors $\mathfrak{P}^j, j \in \{1, 2\}$
are extracted from different layers of DenseNet$201$. The GRB and the FEB are only
conducted on $\mathfrak{P}^1$ with higher resolution while the local patch retrieval is
performed on both of the two tensors. The yielded two anomaly score maps
$\text{A}^{1}_{\text{ori}}$ and $\text{A}^{2}_{\text{ori}}$ are firstly aggregated and
then fused with the foreground estimation $\text{F}^{\ast}$ to generate the final score
map $\text{A}^{\ast}$.
In addition, the image-level anomaly score $\alpha^{\ast}$ is estimated using the similar
method as \cite{zhang2022prototypical, zhang2023destseg}.

\begin{algorithm}[htbp]
  \caption{Inference pseudo-code of CPR}
  \label{alg:infer}
  \LinesNumbered
  \KwIn{Two pre-trained feature extractors of DenseNet$201$: $\Psi_{D}^l(\cdot), l \in
    \{1, 2\}$; GRB $\Psi_{G}(\cdot)$, global feature oracle $\mathcal{G}_{\text{ref}}$;
  LRB $\Psi_{L}(\cdot)$; local feature oracles with multi-scale
  $\mathcal{L}^l_{\text{ref}}, l \in \{1, 2\}$; FEB $\Psi_{F}(\cdot)$; foreground
  estimation map set $\mathcal{F}_{\text{ref}}$; number of nearest neighbors $K$; test
  image $\II_{\text{tst}}$.
  }
  \KwOut{Anomaly score map $\text{A}_{\text{tst}}$ and anomaly score $\alpha_{\text{tst}}$ for $\II_{\text{tst}}$.}  
  \textbf{Algorithm:}\\
  \tcc{{Feature extraction}}
  \For{$l \in \{1, 2\}$}{
  $\mathfrak{P}_{\text{tst}}^l = \Psi_{D}^l(\II_{\text{tst}})$
  }
  $\mathfrak{G}_{\text{tst}}, \text{F}_{\text{tst}} =
    \Psi_{G}(\mathfrak{P}_{\text{tst}}^1),\Psi_{F}(\mathfrak{P}_{\text{tst}}^1)$ \\
  \tcc{{Find the $K$-NNs of $\mathfrak{G}_{\text{tst}}$ in
  $\mathcal{G}_{\text{ref}}$}}
  $\boldsymbol{\beta}_{\text{knn}} = \texttt{KNN-Idx}(\mathfrak{G}_{\text{tst}},
    \mathcal{G}_{\text{ref}}) \in \mathbb{N}^{K}$\\

  $\mathcal{F}_{\text{knn}} = \mathcal{F}_{\text{ref}}[\boldsymbol{\beta}_{\text{knn}}]$\\

  $\text{F}^* = \texttt{MaxPool}(\mathcal{F}_{\text{knn}} \cup \text{F}_{\text{tst}})$ \\
  \tcc{{multiscale local retrieval}}
  $\text{A}_{\text{mul}} = \text{0} \in \mathbb{R}^{H_L \times W_L}$\\

  \For{$l \in \{1, 2\}$}
  {

  $\mathfrak{L}^l_{\text{tst}} = \Psi_{L}(\mathfrak{P}^l_{\text{tst}})$ \\

  $\mathfrak{L}^l_{\text{tst}}
    \xrightarrow{\texttt{flatten}} \mathcal{L}_{\text{tst}} = \{\vv{l}^{r, c}_{\text{tst}} \in \hanxiR^{C_L}
    \mid \forall r, c\}$ \\

  $\mathcal{L}^l_{\text{ref}}[\boldsymbol{\beta}_{\text{knn}}]
    \xrightarrow{\texttt{flatten}} \mathcal{L}_{\text{knn}} = \{\vv{l}^{r, c, j}_{\text{knn}} \in \hanxiR^{C_L}
    \mid \forall r, c, j\}$ \\

  \For{$r \in \{1, \dots, H_L\}$ and $c \in \{1, \dots, W_L\}$}
  {
  $\vv{l}^{\ast}_{r, c} = \texttt{LocalNN}(\vv{l}^{r, c}_{\text{tst}},
    \mathcal{L}_{\text{knn}})$ as \ref{equ:loal_retrieval}
  $\text{A}_{\text{ori}}^l[r,c] = 1 -
    \Psi_{\text{cos}}(\vv{l}^{r, c}_{\text{tst}}, \vv{l}^{\ast}_{r, c})$\\

  }
  $\text{A}_{\text{mul}} = \text{A}_{\text{mul}} +
    \texttt{UpInterp}(\text{A}_{\text{ori}}^l)$\\
  }
  \tcc{{foreground filter}}
  $\text{A}^{\ast} = \text{A}_{\text{mul}} \odot \text{F}^{\ast}$\\
  \tcc{{image-level anomaly score}}
  $\alpha^{\ast} =
    \sum_{i=1}^{T}\texttt{sort}^{\Downarrow}(\texttt{flatten}(\text{A}^{\ast}))$
\end{algorithm}

\subsection{Implementation details}
\label{subsec:cpr_detail}
The input images of our CPR model are consistently resized to $320 \times 320$, in both
training and test procedures.
We adopt the ``denseblock-$1$'' and ``denseblock-$2$'' of the DenseNet$201$ model
\cite{huang2018densely} pre-trained on ImageNet \cite{russakovsky2015imagenet} to obtain
the raw feature tensors with the sizes $256 \times 80 \times 80$ and $512 \times 40 \times
  40$ respectively and freeze these blocks during training. In particular, GRB and FEB only
performed based on denseblock-$1$ and LRB uses both denseblock-$1$ and denseblock-$2$ in a
multi-scale manner.

The dimension of the feature space for local retrieval is $384$, following the setting in
\cite{roth2022towards}. In the inference stage, the number of $K$ nearest neighbors is $10$,
the size of the sub-tensors $S$ is set to 5, the number of clusters $N_c$ is
$12$, the parameter for robust KL divergences $\tau$ is set to $5$, the parameter $T$ for
calculating image-level anomaly score is $512$ and the local retrieval region size
for denseblock-$1$ and denseblock-$2$ are $3$ and $1$ respectively.

As to the training setup, we use the AdamW optimizer \cite{loshchilov2018decoupled} to
update the model parameters with a default learning rate $1 \times 10^{-3}$ and default
weight decay rate $1 \times 10^{-2}$. The training is conducted for $40000$ iterations
with the batch size $32$ for each sub-category and we empirically select \textbf{A FIXED}
iteration number for \textbf{ALL THE SUB-CATEGORIES} of one dataset \footnote{Note that
  for the most comprehensive MVTec AD dataset \cite{Bergmann_2019_CVPR}, we treat it as a
  combination of a texture subset and an object subset.}. In this paper, the data augmentation only contains
the salt-and-pepper noise and the brightness changes.

We adopt off-the-shelf defect generation/augmentation approaches in this paper. For the
anomaly-free training sets, synthetic anomalies are randomly generated and added onto the
normal samples, mainly following the pseudo-defect generation method proposed in
\cite{yang2023memseg}. For the supervised scenarios, we employ a similar strategy as the
PRN algorithm \cite{zhang2022prototypical} to ``extend'' the anomalous patterns for
achieving better performance based on limited samples.

\section{Experiments}
\label{sec:experiments}
In this section, a series of experiments are conducted to verify the proposed method,
comparing with other state-of-the-art methods. Three well-acknowledged benchmark datasets, namely the MVTec AD \cite{bergmann2019mvtec}, MVTec-3D AD \cite{Bergmann_2022} and BTAD
\cite{mishra2021vt} are involved in the experiments. The anomaly detection and
localization performances are evaluated using four popular AD metrics, \emph{i.e.},
Image-AUC, Pixel-AUC, Per-Region Overlap (PRO), and Average Precision (AP), under both the
supervised and unsupervised scenarios.

\subsection{Experimental settings}
\label{subsec:exp_setting}
\subsubsection{Three involved datasets}
\begin{itemize}
  \item
        \textbf{MVTec AD} \cite{Bergmann_2019_CVPR} contains $5354$ images belonging to $15$
        sub-datasets which can be further split into $10$ object sub-categories and $5$
        texture sub-categories. Each sub-category involves a nominal-only training set and a
        test set with various types of anomalies.
  \item
        \textbf{MVTec-3D AD} \cite{Bergmann_2022} is a 3D anomaly detection dataset
        comprising over $4000$ RGB images and the corresponding high-resolution 3D point
        cloud data. Similar to MVTec AD, each of the $10$ sub-categories of MVTec-3D AD are
        divided into a defect-free training set and a test set containing various kinds of
        defects. Though MVTec-3D AD provides accurate point cloud information, only RGB
        images are utilized for predicting anomalies in this work.
  \item
        \textbf{BTAD} \cite{mishra2021vt} comprises $2540$ images from three categories of
        real-world industrial products with different body and surface defects. It is
        usually considered as a complementary dataset to the MVTec AD when evaluating an AD
        algorithm.
\end{itemize}

\subsubsection{Four evaluation metrics}
\begin{itemize}
  \item
        \textbf{Image-AUC}: this image-level AD performance is measured by calculating the
        area under the ``Receiver Operating Characteristic'' curve (AUROC) of the predicted
        image-level anomaly scores with increasing thresholds.
  \item
        \textbf{Pixel-AUC}: this pixel-level AD performance is similar to the image-AUC
        except that the unit to be classified is pixel rather than image.
  \item
        \textbf{PRO}: the ``Per-Region Overlap'' metric is a pixel-level metric that evaluates
        the AD performance on connected anomaly regions. It is more robust to the dataset
        with significantly different sizes of anomaly regions \cite{bergmann2020uninformed}.
  \item
        \textbf{AP}: the average precision (AP) is the most popular metric for semantic
        segmentation tasks. It treats each pixel independently when measuring the
        segmentation performance \cite{saito2015precision}.
\end{itemize}

\subsubsection{Two supervision scenarios}
\begin{itemize}
  \item
        \textbf{The unsupervised scenario.} This is the conventional supervision condition
        in the literature of AD \cite{Bergmann_2019_CVPR, Bergmann_2022, mishra2021vt}. In
        this scenario, \textbf{all} the training images are anomaly free while various
        anomalies exist in the test set. To achieve a better AD model, we train our CPR
        algorithm by using synthetic defects as it is introduced in \Cref{subsec:cpr_learn}.
  \item
        \textbf{The supervised scenario.} To mimic the real-life situation where a limited
        number of defective samples are available, some researchers \cite{ding2022catching,
          zhang2022prototypical} propose to randomly select some anomalous test samples and
        add them to the original training set. The yielded AD models usually perform better
        on the remaining test images. In this work, we employ the ``Extended Anomalies''
        strategy \cite{zhang2022prototypical} to further exploit the information of the
        genuine defects.
\end{itemize}

\subsubsection{Hardware setting}
We conduct all the mentioned experiments in this paper using a machine equipped with an
Intel i$5$-$13600$KF CPU, 32G DDR4 RAM and an NVIDIA RTX $4090$ GPU.

\subsection{Quantitative Results}
\label{subsec:quantitative}

\subsubsection{Results in unsupervised setting}
We firstly evaluate the proposed method and the $8$ comparing SOTA methods (PatchCore
\cite{roth2022towards}, DRAEM \cite{zavrtanik2021draem}, RD \cite{deng2022anomaly}, SSPCAB
\cite{ristea2022self}, DeSTSeg \cite{zhang2023destseg}, PyramidFlow
\cite{lei2023pyramidflow}, CDO \cite{cao2023collaborative} and SemiREST \cite{li2023efficient}) on the
MVTec AD dataset \cite{Bergmann_2019_CVPR} and within the unsupervised condition.
\Cref{table:result} shows the AP, PRO and Pixel-AUC results of the involved methods.
According to the table, our CPR algorithm outperforms other competitors by large margins.
The most recently proposed SemiREST algorithm \cite{li2023efficient} performs slightly
better than our method on the texture sub-categories while CPR beats it on the object
sub-categories with significant superiority (\emph{e.g.}, the nearly $3\%$ gain on AP). In
general, the proposed CPR method ranks first for the AP and PRO metrics and ranks
second for the Pixel-AUC metric. We report the Image-AUC scores of the comparing methods
in \Cref{table:image_auc_result}, from which one can see a new Image-AUC record of
$99.8\%$ is achieved by the proposed method. This performance is achieved by selecting the
best training iteration number independently for each sub-category. Although contradicting
the principle of model selection described in \Cref{subsec:cpr_detail}, this model
selection trick is common in the literature of anomaly detection
\cite{gudovskiy2022cflow, Liu_2023_CVPR, cao2023collaborative}. When strictly following the model
selection principle in \Cref{subsec:cpr_detail}, the Image-AUC drops marginally to
$99.7\%$ which is also higher than the existing SOTA performances.

\begin{table*}[htbp]
  \centering
  \caption{The AP, PRO and Pixel-AUROC scores on MVTec AD in the unsupervised settings.}
  \label{table:result}
  \resizebox{\textwidth}{!}{
    \begin{tabular}{lccccccccc}
      \toprule
      Category     & \makecell{PatchCore \cite{roth2022towards} \NL (CVPR2022)} & \makecell{DRAEM \cite{zavrtanik2021draem} \NL (ICCV2021)}                                     & \makecell{RD \cite{deng2022anomaly} \NL (CVPR2022)}                   & \makecell{SSPCAB \cite{ristea2022self} \NL (CVPR2022)}                                        & \makecell{DeSTSeg \cite{zhang2023destseg} \NL (CVPR2023)}            & \makecell{PyramidFlow \cite{lei2023pyramidflow} \NL (CVPR2023)} & \makecell{CDO \cite{cao2023collaborative} \NL (TII2023)} & \makecell{SemiREST \cite{li2023efficient} \NL (arXiv2023)}                                      & Ours                                                                                            \\\hline
      Carpet       & $64.1$/$95.1$/{\color{blue}{$\mathbf{99.1}$}}              & $53.5$/$92.9$/$95.5$                                                                          & $56.5$/$95.4$/$98.9$                                                  & $48.6$/$86.4$/$92.6$                                                                          & $72.8$/$93.6$/$96.1$                                                 & $\sim$/$97.2$/$97.4$                                            & $53.4$/$96.8$/{\color{blue}{$\mathbf{99.1}$}}            & {\color{red}{$\mathbf{84.2}$}}/{\color{red}{$\mathbf{98.7}$}}/{\color{red}{$\mathbf{99.6}$}}    & {\color{blue}{$\mathbf{81.2}$}}/{\color{blue}{$\mathbf{97.6}$}}/$98.9$                          \\
      Grid         & $30.9$/$93.6$/$98.8$                                       & {\color{red}{$\mathbf{65.7}$}}/{\color{red}{$\mathbf{98.3}$}}/{\color{red}{$\mathbf{99.7}$}}  & $15.8$/$94.2$/$98.3$                                                  & $57.9$/{\color{blue}{$\mathbf{98.0}$}}/{\color{blue}{$\mathbf{99.5}$}}                        & $61.5$/$96.4$/$99.1$                                                 & $\sim$/$94.3$/$95.7$                                            & $45.3$/$96.1$/$98.4$                                     & {\color{blue}{$\mathbf{65.5}$}}/$97.9$/{\color{blue}{$\mathbf{99.5}$}}                          & $64.0$/$97.6$/{\color{blue}{$\mathbf{99.5}$}}                                                   \\
      Leather      & $45.9$/$97.2$/$99.3$                                       & $75.3$/$97.4$/$98.6$                                                                          & $47.6$/$98.2$/$99.4$                                                  & $60.7$/$94.0$/$96.3$                                                                          & $75.6$/$99.0$/{\color{blue}{$\mathbf{99.7}$}}                        & $\sim$/$99.2$/$98.7$                                            & $43.6$/$98.3$/$99.2$                                     & {\color{red}{$\mathbf{79.3}$}}/{\color{blue}{$\mathbf{99.4}$}}/{\color{red}{$\mathbf{99.8}$}}   & {\color{blue}{$\mathbf{78.5}$}}/{\color{red}{$\mathbf{99.6}$}}/{\color{red}{$\mathbf{99.8}$}}   \\
      Tile         & $54.9$/$80.2$/$95.7$                                       & $92.3$/{\color{blue}{$\mathbf{98.2}$}}/$99.2$                                                 & $54.1$/$85.6$/$95.7$                                                  & {\color{blue}{$\mathbf{96.1}$}}/$98.1$/{\color{blue}{$\mathbf{99.4}$}}                        & $90.0$/$95.5$/$98.0$                                                 & $\sim$/$97.2$/$97.1$                                            & $61.8$/$90.5$/$97.2$                                     & {\color{red}{$\mathbf{96.4}$}}/{\color{red}{$\mathbf{98.5}$}}/{\color{red}{$\mathbf{99.7}$}}    & $94.1$/$98.1$/$99.2$                                                                            \\
      Wood         & $50.0$/$88.3$/$95.0$                                       & $77.7$/$90.3$/$96.4$                                                                          & $48.3$/$91.4$/$95.8$                                                  & $78.9$/$92.8$/$96.5$                                                                          & {\color{red}{$\mathbf{81.9}$}}/$96.1$/{\color{red}{$\mathbf{97.7}$}} & $\sim$/{\color{red}{$\mathbf{97.9}$}}/$97.0$                    & $46.3$/$92.9$/$95.8$                                     & $79.4$/$96.5$/{\color{red}{$\mathbf{97.7}$}}                                                    & {\color{blue}{$\mathbf{80.8}$}}/{\color{blue}{$\mathbf{97.7}$}}/{\color{blue}{$\mathbf{97.4}$}} \\\hline
      Average      & $49.2$/$90.9$/$97.6$                                       & $72.9$/$95.4$/$97.9$                                                                          & $44.5$/$93.0$/$97.6$                                                  & $68.4$/$93.9$/$96.9$                                                                          & $76.4$/$96.1$/$98.1$                                                 & $\sim$/{\color{blue}{$\mathbf{97.2}$}}/$97.2$                   & $50.1$/$96.5$/$98.0$                                     & {\color{red}{$\mathbf{81.0}$}}/{\color{red}{$\mathbf{98.2}$}}/{\color{red}{$\mathbf{99.3}$}}    & {\color{blue}{$\mathbf{79.7}$}}/{\color{red}{$\mathbf{98.2}$}}/{\color{blue}{$\mathbf{99.0}$}}  \\\hline
      Bottle       & $77.7$/$94.7$/$98.5$                                       & $86.5$/$96.8$/$99.1$                                                                          & $78.0$/$96.3$/$98.8$                                                  & $89.4$/$96.3$/$99.2$                                                                          & $90.3$/$96.6$/$99.2$                                                 & $\sim$/$95.5$/$97.8$                                            & $84.1$/$97.2$/$99.3$                                     & {\color{red}{$\mathbf{94.1}$}}/{\color{red}{$\mathbf{98.6}$}}/{\color{red}{$\mathbf{99.6}$}}    & {\color{blue}{$\mathbf{92.6}$}}/{\color{blue}{$\mathbf{98.1}$}}/{\color{blue}{$\mathbf{99.4}$}} \\
      Cable        & $66.3$/$93.2$/$98.4$                                       & $52.4$/$81.0$/$94.7$                                                                          & $52.6$/$94.1$/$97.2$                                                  & $52.0$/$80.4$/$95.1$                                                                          & $60.4$/$86.4$/$97.3$                                                 & $\sim$/$90.3$/$91.8$                                            & $61.0$/$94.2$/$97.6$                                     & {\color{blue}{$\mathbf{81.1}$}}/{\color{red}{$\mathbf{95.3}$}}/{\color{blue}{$\mathbf{99.1}$}}  & {\color{red}{$\mathbf{84.4}$}}/{\color{blue}{$\mathbf{95.2}$}}/{\color{red}{$\mathbf{99.3}$}}   \\
      Capsule      & $44.7$/$94.8$/$99.0$                                       & $49.4$/$82.7$/$94.3$                                                                          & $47.2$/$95.5$/$98.7$                                                  & $46.4$/$92.5$/$90.2$                                                                          & $56.3$/$94.2$/{\color{blue}{$\mathbf{99.1}$}}                        & $\sim$/{\color{red}{$\mathbf{98.3}$}}/$98.6$                    & $39.5$/$93.0$/$98.6$                                     & {\color{blue}{$\mathbf{57.2}$}}/{\color{blue}{$\mathbf{96.9}$}}/$98.8$                          & {\color{red}{$\mathbf{60.4}$}}/$96.3$/{\color{red}{$\mathbf{99.3}$}}                            \\
      Hazelnut     & $53.5$/$95.2$/$98.7$                                       & {\color{blue}{$\mathbf{92.9}$}}/{\color{red}{$\mathbf{98.5}$}}/{\color{red}{$\mathbf{99.7}$}} & $60.7$/$96.9$/$99.0$                                                  & {\color{red}{$\mathbf{93.4}$}}/{\color{blue}{$\mathbf{98.2}$}}/{\color{red}{$\mathbf{99.7}$}} & $88.4$/$97.6$/{\color{blue}{$\mathbf{99.6}$}}                        & $\sim$/$98.1$/$98.1$                                            & $66.1$/$97.4$/$99.2$                                     & $87.8$/$96.1$/{\color{blue}{$\mathbf{99.6}$}}                                                   & $88.7$/$97.6$/{\color{blue}{$\mathbf{99.6}$}}                                                   \\
      Metal Nut    & $86.9$/$94.0$/$98.3$                                       & {\color{blue}{$\mathbf{96.3}$}}/$97.0$/{\color{red}{$\mathbf{99.5}$}}                         & $78.6$/$94.9$/$97.3$                                                  & $94.7$/{\color{red}{$\mathbf{97.7}$}}/{\color{blue}{$\mathbf{99.4}$}}                         & $93.5$/$95.0$/$98.6$                                                 & $\sim$/$91.4$/$97.2$                                            & $83.8$/$95.7$/$98.5$                                     & {\color{red}{$\mathbf{96.6}$}}/{\color{blue}{$\mathbf{97.5}$}}/{\color{red}{$\mathbf{99.5}$}}   & $93.5$/{\color{blue}{$\mathbf{97.5}$}}/$99.3$                                                   \\
      Pill         & $77.9$/$95.0$/$97.8$                                       & $48.5$/$88.4$/$97.6$                                                                          & $76.5$/$96.7$/$98.1$                                                  & $48.3$/$89.6$/$97.2$                                                                          & $83.1$/$95.3$/$98.7$                                                 & $\sim$/$96.1$/$96.1$                                            & $81.1$/$96.6$/$98.9$                                     & {\color{blue}{$\mathbf{85.9}$}}/{\color{blue}{$\mathbf{98.4}$}}/{\color{blue}{$\mathbf{99.2}$}} & {\color{red}{$\mathbf{91.5}$}}/{\color{red}{$\mathbf{98.7}$}}/{\color{red}{$\mathbf{99.5}$}}    \\
      Screw        & $36.1$/$97.1$/{\color{blue}{$\mathbf{99.5}$}}              & $58.2$/$95.0$/$97.6$                                                                          & $52.1$/{\color{blue}{$\mathbf{98.5}$}}/{\color{red}{$\mathbf{99.7}$}} & $61.7$/$95.2$/$99.0$                                                                          & $58.7$/$92.5$/$98.5$                                                 & $\sim$/$94.7$/$94.6$                                            & $39.4$/$94.3$/$99.0$                                     & {\color{blue}{$\mathbf{65.9}$}}/$97.9$/{\color{red}{$\mathbf{99.7}$}}                           & {\color{red}{$\mathbf{71.0}$}}/{\color{red}{$\mathbf{98.7}$}}/{\color{red}{$\mathbf{99.7}$}}    \\
      Toothbrush   & $38.3$/$89.4$/$98.6$                                       & $44.7$/$85.6$/$98.1$                                                                          & $51.1$/$92.3$/$99.1$                                                  & $39.3$/$85.5$/$97.3$                                                                          & {\color{blue}{$\mathbf{75.2}$}}/$94.0$/$99.3$                        & $\sim$/{\color{blue}{$\mathbf{97.9}$}}/$98.5$                   & $45.9$/$90.5$/$98.9$                                     & $74.5$/$96.2$/{\color{blue}{$\mathbf{99.5}$}}                                                   & {\color{red}{$\mathbf{84.1}$}}/{\color{red}{$\mathbf{98.0}$}}/{\color{red}{$\mathbf{99.7}$}}    \\
      Transistor   & $66.4$/$92.4$/$96.3$                                       & $50.7$/$70.4$/$90.9$                                                                          & $54.1$/$83.3$/$92.3$                                                  & $38.1$/$62.5$/$84.8$                                                                          & $64.8$/$85.7$/$89.1$                                                 & $\sim$/$94.7$/{\color{blue}{$\mathbf{96.9}$}}                   & $56.3$/$92.6$/$95.3$                                     & {\color{blue}{$\mathbf{79.4}$}}/{\color{blue}{$\mathbf{96.0}$}}/{\color{red}{$\mathbf{98.0}$}}  & {\color{red}{$\mathbf{86.7}$}}/{\color{red}{$\mathbf{97.1}$}}/{\color{red}{$\mathbf{98.0}$}}    \\
      Zipper       & $62.8$/$95.8$/$98.9$                                       & $81.5$/$96.8$/$98.8$                                                                          & $57.5$/$95.3$/$98.3$                                                  & $76.4$/$95.2$/$98.4$                                                                          & $85.2$/$97.4$/$99.1$                                                 & $\sim$/$95.4$/$96.6$                                            & $55.6$/$94.3$/$98.2$                                     & {\color{red}{$\mathbf{90.2}$}}/{\color{red}{$\mathbf{98.9}$}}/{\color{red}{$\mathbf{99.7}$}}    & {\color{blue}{$\mathbf{88.8}$}}/{\color{blue}{$\mathbf{98.6}$}}/{\color{blue}{$\mathbf{99.6}$}} \\\hline
      Average      & $61.1$/$94.2$/$98.4$                                       & $66.1$/$89.2$/$97.0$                                                                          & $60.8$/$94.4$/$97.9$                                                  & $64.0$/$89.3$/$96.0$                                                                          & $75.6$/$93.5$/$97.5$                                                 & $\sim$/$95.2$/$96.6$                                            & $61.3$/$94.6$/$98.4$                                     & {\color{blue}{$\mathbf{81.3}$}}/{\color{blue}{$\mathbf{97.2}$}}/{\color{blue}{$\mathbf{99.3}$}} & {\color{red}{$\mathbf{84.2}$}}/{\color{red}{$\mathbf{97.6}$}}/{\color{red}{$\mathbf{99.4}$}}    \\\hline
      TotalAverage & $57.1$/$93.1$/$98.1$                                       & $68.4$/$91.3$/$97.3$                                                                          & $55.4$/$93.9$/$97.8$                                                  & $65.5$/$90.8$/$96.3$                                                                          & $75.8$/$94.4$/$97.9$                                                 & $\sim$/$95.9$/$96.8$                                            & $57.6$/$94.7$/$98.2$                                     & {\color{blue}{$\mathbf{81.2}$}}/{\color{blue}{$\mathbf{97.5}$}}/{\color{red}{$\mathbf{99.3}$}}  & {\color{red}{$\mathbf{82.7}$}}/{\color{red}{$\mathbf{97.8}$}}/{\color{blue}{$\mathbf{99.2}$}}   \\
      \bottomrule
    \end{tabular}
  }
\end{table*}

\begin{table*}[htbp]
  \centering
  \caption{
    Image-level anomaly detection AUC (\%) on MVTec AD. Results are averaged over
    all categories.
  }
  \label{table:image_auc_result}
  \begin{tabular}{ccccccccc}
    \toprule
    \makecell{PatchCore \cite{roth2022towards} \NL (CVPR2022)} & \makecell{DRAEM \cite{zavrtanik2021draem} \NL (ICCV2021)} & \makecell{RD \cite{deng2022anomaly} \NL (CVPR2022)} & \makecell{SSPCAB \cite{ristea2022self} \NL (CVPR2022)} & \makecell{DeSTSeg \cite{zhang2023destseg} \NL (CVPR2023)} & \makecell{CDO \cite{cao2023collaborative} \NL (TII2023)} & \makecell{SimpleNet \cite{Liu_2023_CVPR} \NL (CVPR2023)} & Ours                            & Ours$^*$                       \\ \hline
    $98.5$                                                     & $98.0$                                                    & $98.5$                                              & $98.9$                                                 & $98.6$                                                    & $96.8$                                                   & $99.6$                                                   & {\color{blue}{$\mathbf{99.7}$}} & {\color{red}{$\mathbf{99.8}$}} \\
    \bottomrule
  \end{tabular}
\end{table*}

To achieve a more comprehensive comparison, the proposed method is also evaluated on other
two widely-used datasets, namely MVTec-3D AD \cite{Bergmann_2022} and BTAD
\cite{mishra2021vt}. The corresponding results of CPR and those SOTA algorithms on these
two datasets are shown in \Cref{table:3d_result} and \Cref{table:btad_supervised_result}.
Unsurprisingly, the proposed CPR maintains the accuracy superiority over the existing SOTA
methods.

\begin{table}[htbp]
  \centering
  \caption{
    AP, PRO and Pixel-AUROC scores on MVTec-3D
    AD \cite{Bergmann_2022} with pure RGB inputs.
  }
  \label{table:3d_result}
  \resizebox{\linewidth}{!}{
    \begin{tabular}{lcccc}
      \toprule
      Category   & \makecell{AST \cite{rudolph2022asymmetric} \NL (WACV2023)} & \makecell{CDO \cite{cao2023collaborative} \NL (TII2023)}                                       & \makecell{M3DM \cite{wang2023multimodal} \NL (CVPR2023)}                                        & Ours                                                                                         \\\hline
      Bagel      & $20.5$/$81.0$/$95.9$                                       & $50.4$/{\color{blue}{$\mathbf{98.0}$}}/{\color{blue}{$\mathbf{99.3}$}}                         & {\color{blue}{$\mathbf{58.1}$}}/$94.5$/$99.1$                                                   & {\color{red}{$\mathbf{83.3}$}}/{\color{red}{$\mathbf{99.5}$}}/{\color{red}{$\mathbf{99.8}$}} \\
      CableGland & $11.5$/$87.0$/$94.7$                                       & {\color{blue}{$\mathbf{42.7}$}}/{\color{red}{$\mathbf{98.5}$}}/{\color{blue}{$\mathbf{99.4}$}} & $40.6$/{\color{blue}{$\mathbf{97.6}$}}/{\color{blue}{$\mathbf{99.4}$}}                          & {\color{red}{$\mathbf{61.5}$}}/{\color{red}{$\mathbf{98.5}$}}/{\color{red}{$\mathbf{99.6}$}} \\
      Carrot     & $7.6$/$81.5$/$90.8$                                        & $27.5$/{\color{red}{$\mathbf{97.9}$}}/{\color{red}{$\mathbf{99.4}$}}                           & {\color{blue}{$\mathbf{32.1}$}}/{\color{blue}{$\mathbf{97.3}$}}/{\color{red}{$\mathbf{99.4}$}}  & {\color{red}{$\mathbf{37.5}$}}/$96.8$/{\color{blue}{$\mathbf{99.0}$}}                        \\
      Cookie     & $18.4$/$46.6$/$81.8$                                       & $49.9$/{\color{blue}{$\mathbf{88.7}$}}/{\color{blue}{$\mathbf{98.0}$}}                         & {\color{blue}{$\mathbf{50.9}$}}/$88.5$/$97.1$                                                   & {\color{red}{$\mathbf{59.8}$}}/{\color{red}{$\mathbf{94.6}$}}/{\color{red}{$\mathbf{98.3}$}} \\
      Dowel      & $12.6$/$92.5$/$98.2$                                       & $44.3$/$97.5$/{\color{blue}{$\mathbf{99.6}$}}                                                  & {\color{blue}{$\mathbf{51.3}$}}/{\color{blue}{$\mathbf{97.6}$}}/{\color{red}{$\mathbf{99.7}$}}  & {\color{red}{$\mathbf{58.6}$}}/{\color{red}{$\mathbf{98.5}$}}/{\color{red}{$\mathbf{99.7}$}} \\
      Foam       & $5.5$/$54.8$/$72.7$                                        & $20.5$/$68.1$/$87.6$                                                                           & {\color{blue}{$\mathbf{33.0}$}}/{\color{blue}{$\mathbf{84.5}$}}/{\color{blue}{$\mathbf{95.6}$}} & {\color{red}{$\mathbf{52.7}$}}/{\color{red}{$\mathbf{90.8}$}}/{\color{red}{$\mathbf{97.3}$}} \\
      Peach      & $12.6$/$85.2$/$93.5$                                       & {\color{blue}{$\mathbf{51.2}$}}/{\color{red}{$\mathbf{98.6}$}}/{\color{blue}{$\mathbf{99.6}$}} & $44.3$/{\color{blue}{$\mathbf{97.0}$}}/$99.4$                                                   & {\color{red}{$\mathbf{65.0}$}}/{\color{red}{$\mathbf{98.6}$}}/{\color{red}{$\mathbf{99.7}$}} \\
      Potato     & $3.1$1/$52.6$/$69.4$                                       & $18.2$/{\color{red}{$\mathbf{95.3}$}}/{\color{red}{$\mathbf{99.1}$}}                           & {\color{blue}{$\mathbf{24.7}$}}/{\color{red}{$\mathbf{95.3}$}}/{\color{blue}{$\mathbf{99.0}$}}  & {\color{red}{$\mathbf{28.4}$}}/{\color{blue}{$\mathbf{95.0}$}}/$98.6$                        \\
      Rope       & $17.5$/$86.8$/$96.8$                                       & $41.1$/{\color{blue}{$\mathbf{96.8}$}}/{\color{blue}{$\mathbf{99.4}$}}                         & {\color{blue}{$\mathbf{50.8}$}}/$94.9$/$99.3$                                                   & {\color{red}{$\mathbf{74.8}$}}/{\color{red}{$\mathbf{98.3}$}}/{\color{red}{$\mathbf{99.7}$}} \\
      Tire       & $4.3$/$54.8$/$83.3$                                        & $36.7$/{\color{blue}{$\mathbf{97.8}$}}/{\color{blue}{$\mathbf{99.5}$}}                         & {\color{blue}{$\mathbf{40.6}$}}/$97.1$/{\color{blue}{$\mathbf{99.5}$}}                          & {\color{red}{$\mathbf{55.8}$}}/{\color{red}{$\mathbf{98.6}$}}/{\color{red}{$\mathbf{99.7}$}} \\\hline
      Average    & $11.4$/$72.3$/$87.7$                                       & $38.2$/$93.7$/$98.1$                                                                           & {\color{blue}{$\mathbf{42.6}$}}/{\color{blue}{$\mathbf{94.4}$}}/{\color{blue}{$\mathbf{98.7}$}} & {\color{red}{$\mathbf{57.8}$}}/{\color{red}{$\mathbf{96.9}$}}/{\color{red}{$\mathbf{99.1}$}} \\
      \bottomrule
    \end{tabular}
  }
\end{table}

\begin{table*}[htbp]
  \centering
  \caption{
    Results of the AP, PRO and Pixel-AUC metrics for unsupervised anomaly localization
    performance on BTAD \cite{mishra2021vt}.
  }
  \label{table:btad_result}
  \begin{tabular}{lcccccc}
    \toprule
    Category & \makecell{PatchCore \cite{roth2022towards} \NL (CVPR2022)} & \makecell{SSPCAB \cite{ristea2022self} \NL (CVPR2022)} & \makecell{RD \cite{deng2022anomaly} \NL (CVPR2022)} & \makecell{NFAD \cite{yao2022explicit} \NL (CVPR2023)}                                        & \makecell{SemiREST \cite{li2023efficient} \NL (arXiv2023)}                                      & Ours                                                                                           \\\hline
    $01$     & $47.1$/$78.4$/$96.5$                                       & $18.1$/$62.8$/$92.4$                                   & $49.3$/$72.8$/$95.7$                                & $46.7$/$76.6$/$96.7$                                                                         & {\color{blue}{$\mathbf{52.4}$}}/{\color{blue}{$\mathbf{83.9}$}}/{\color{blue}{$\mathbf{97.5}$}} & {\color{red}{$\mathbf{72.4}$}}/{\color{red}{$\mathbf{88.4}$}}/{\color{red}{$\mathbf{98.5}$}}   \\
    $02$     & $56.3$/$54.0$/$94.9$                                       & $15.8$/$28.6$/$65.6$                                   & {\color{blue}{$\mathbf{66.1}$}}/$55.8$/$96.0$       & $59.2$/$57.9$/$96.4$                                                                         & $63.1$/{\color{blue}{$\mathbf{61.5}$}}/{\color{blue}{$\mathbf{96.5}$}}                          & {\color{red}{$\mathbf{81.6}$}}/{\color{red}{$\mathbf{68.6}$}}/{\color{red}{$\mathbf{97.0}$}}   \\
    $03$     & $51.2$/$96.4$/{\color{blue}{$\mathbf{99.2}$}}              & $5.0$/$71.0$/$92.4$                                    & $45.1$/{\color{red}{$\mathbf{98.8}$}}/$99.0$        & {\color{red}{$\mathbf{62.8}$}}/{\color{red}{$\mathbf{98.8}$}}/{\color{red}{$\mathbf{99.7}$}} & $50.9$/{\color{red}{$\mathbf{98.8}$}}/{\color{red}{$\mathbf{99.7}$}}                            & {\color{blue}{$\mathbf{57.0}$}}/{\color{blue}{$\mathbf{98.2}$}}/{\color{red}{$\mathbf{99.7}$}} \\\hline
    Average  & $51.5$/$76.3$/$96.9$                                       & $13.0$/$54.1$/$83.5$                                   & $53.5$/$75.8$/$96.9$                                & {\color{blue}{$\mathbf{56.2}$}}/$77.8$/$97.6$                                                & $55.5$/{\color{blue}{$\mathbf{81.4}$}}/{\color{blue}{$\mathbf{97.9}$}}                          & {\color{red}{$\mathbf{70.3}$}}/{\color{red}{$\mathbf{85.1}$}}/{\color{red}{$\mathbf{98.4}$}}   \\
    \bottomrule
  \end{tabular}
\end{table*}

\subsubsection{Results in supervised setting}

On the other hand, the proposed CPR method also performs well in the supervised scenario.
According to the results reported in \Cref{table:supervise_result} and
\Cref{table:btad_supervised_result}, the CPR method achieves the consistently highest
scores with Pixel-AUC, PRO and AP metrics. In particular, the overall AP of the proposed
algorithm is $86.0\%$ on MVTec AD and $84.0\%$ on BTAD, exceeding the current SOTA
method SemiREST \cite{li2023efficient} by $1.6\%$ and $2.0\%$, respectively.

\begin{table*}[htbp]
  \centering
  \caption{
    Results of the AP, PRO and Pixel-AUROC metrics for anomaly localization performance on
    MVTec AD in a supervised setting. In accordance with the general framework proposed by
    DRA \cite{ding2022catching}, we sampled $10$ labeled anomaly samples from all anomaly
    samples in the test set for each sub-dataset.
  }
  \label{table:supervise_result}
  \begin{tabular}{lccccc}
    \toprule
    Category     & \makecell{DevNet \cite{pang2021explainable} \NL (arXiv2021)} & \makecell{DRA \cite{ding2022catching} \NL (CVPR2022)} & \makecell{PRN \cite{zhang2022prototypical} \NL (CVPR2023)}                                    & \makecell{SemiREST \cite{li2023efficient} \NL (arXiv2023)}                                      & Ours                                                                                            \\\hline
    Carpet       & $45.7$/$85.8$/$97.2$                                         & $52.3$/$92.2$/$98.2$                                  & $82.0$/$97.0$/$99.0$                                                                          & {\color{red}{$\mathbf{89.1}$}}/{\color{red}{$\mathbf{99.1}$}}/{\color{red}{$\mathbf{99.7}$}}    & {\color{blue}{$\mathbf{88.1}$}}/{\color{blue}{$\mathbf{98.9}$}}/{\color{blue}{$\mathbf{99.6}$}} \\
    Grid         & $25.5$/$79.8$/$87.9$                                         & $26.8$/$71.5$/$86.0$                                  & $45.7$/$95.9$/$98.4$                                                                          & {\color{blue}{$\mathbf{66.4}$}}/{\color{blue}{$\mathbf{97.0}$}}/{\color{blue}{$\mathbf{99.4}$}} & {\color{red}{$\mathbf{67.3}$}}/{\color{red}{$\mathbf{98.7}$}}/{\color{red}{$\mathbf{99.7}$}}    \\
    Leather      & $8.1$/$88.5$/$94.2$                                          & $5.6$/$84.0$/$93.8$                                   & $69.7$/$99.2$/$99.7$                                                                          & {\color{red}{$\mathbf{81.7}$}}/{\color{red}{$\mathbf{99.7}$}}/{\color{red}{$\mathbf{99.9}$}}    & {\color{blue}{$\mathbf{78.0}$}}/{\color{blue}{$\mathbf{99.5}$}}/{\color{blue}{$\mathbf{99.8}$}} \\
    Tile         & $52.3$/$78.9$/$92.7$                                         & $57.6$/$81.5$/$92.3$                                  & $96.5$/$98.2$/{\color{blue}{$\mathbf{99.6}$}}                                                 & {\color{blue}{$\mathbf{96.9}$}}/{\color{blue}{$\mathbf{98.9}$}}/{\color{red}{$\mathbf{99.7}$}}  & {\color{red}{$\mathbf{97.2}$}}/{\color{red}{$\mathbf{99.0}$}}/{\color{red}{$\mathbf{99.7}$}}    \\
    Wood         & $25.1$/$75.4$/$86.4$                                         & $22.7$/$69.7$/$82.9$                                  & $82.6$/$95.9$/$97.8$                                                                          & {\color{blue}{$\mathbf{88.7}$}}/{\color{blue}{$\mathbf{97.9}$}}/{\color{blue}{$\mathbf{99.2}$}} & {\color{red}{$\mathbf{90.7}$}}/{\color{red}{$\mathbf{98.4}$}}/{\color{red}{$\mathbf{99.5}$}}    \\\hline
    Average      & $31.3$/$81.7$/$91.7$                                         & $33.0$/$79.8$/$90.6$                                  & $75.3$/$97.2$/$98.9$                                                                          & {\color{red}{$\mathbf{84.7}$}}/{\color{blue}{$\mathbf{98.5}$}}/{\color{blue}{$\mathbf{99.5}$}}  & {\color{blue}{$\mathbf{84.3}$}}/{\color{red}{$\mathbf{98.9}$}}/{\color{red}{$\mathbf{99.6}$}}   \\\hline
    Bottle       & $51.5$/$83.5$/$93.9$                                         & $41.2$/$77.6$/$91.3$                                  & {\color{blue}{$\mathbf{92.3}$}}/{\color{blue}{$\mathbf{97.0}$}}/$99.4$                        & {\color{red}{$\mathbf{93.6}$}}/{\color{red}{$\mathbf{98.5}$}}/{\color{blue}{$\mathbf{99.5}$}}   & {\color{red}{$\mathbf{93.6}$}}/{\color{red}{$\mathbf{98.5}$}}/{\color{red}{$\mathbf{99.6}$}}    \\
    Cable        & $36.0$/$80.9$/$88.8$                                         & $34.7$/$77.7$/$86.6$                                  & $78.9$/{\color{red}{$\mathbf{97.2}$}}/$98.8$                                                  & {\color{red}{$\mathbf{89.5}$}}/{\color{blue}{$\mathbf{95.9}$}}/{\color{blue}{$\mathbf{99.2}$}}  & {\color{blue}{$\mathbf{88.1}$}}/$94.5$/{\color{red}{$\mathbf{99.4}$}}                           \\
    Capsule      & $15.5$/$83.6$/$91.8$                                         & $11.7$/$79.1$/$89.3$                                  & {\color{blue}{$\mathbf{62.2}$}}/$92.5$/$98.5$                                                 & $60.0$/{\color{red}{$\mathbf{97.0}$}}/{\color{blue}{$\mathbf{98.8}$}}                           & {\color{red}{$\mathbf{65.8}$}}/{\color{blue}{$\mathbf{96.7}$}}/{\color{red}{$\mathbf{99.4}$}}   \\
    Hazelnut     & $22.1$/$83.6$/$91.1$                                         & $22.5$/$86.9$/$89.6$                                  & {\color{blue}{$\mathbf{93.8}$}}/$97.4$/{\color{blue}{$\mathbf{99.7}$}}                        & $92.2$/{\color{blue}{$\mathbf{98.3}$}}/{\color{red}{$\mathbf{99.8}$}}                           & {\color{red}{$\mathbf{94.4}$}}/{\color{red}{$\mathbf{98.7}$}}/{\color{red}{$\mathbf{99.8}$}}    \\
    Metal Nut    & $35.6$/$76.9$/$77.8$                                         & $29.9$/$76.7$/$79.5$                                  & $98.0$/$95.8$/$99.7$                                                                          & {\color{red}{$\mathbf{99.1}$}}/{\color{blue}{$\mathbf{98.2}$}}/{\color{red}{$\mathbf{99.9}$}}   & {\color{blue}{$\mathbf{98.6}$}}/{\color{red}{$\mathbf{98.4}$}}/{\color{blue}{$\mathbf{99.8}$}}  \\
    Pill         & $14.6$/$69.2$/$82.6$                                         & $21.6$/$77.0$/$84.5$                                  & {\color{red}{$\mathbf{91.3}$}}/{\color{blue}{$\mathbf{97.2}$}}/{\color{red}{$\mathbf{99.5}$}} & $86.1$/{\color{red}{$\mathbf{98.9}$}}/{\color{blue}{$\mathbf{99.3}$}}                           & {\color{blue}{$\mathbf{90.7}$}}/{\color{red}{$\mathbf{98.9}$}}/{\color{red}{$\mathbf{99.5}$}}   \\
    Screw        & $1.4$/$31.1$/$60.3$                                          & $5.0$/$30.1$/$54.0$                                   & $44.9$/$92.4$/{\color{blue}{$\mathbf{97.5}$}}                                                 & {\color{blue}{$\mathbf{72.1}$}}/{\color{blue}{$\mathbf{98.8}$}}/{\color{red}{$\mathbf{99.8}$}}  & {\color{red}{$\mathbf{72.5}$}}/{\color{red}{$\mathbf{98.9}$}}/{\color{red}{$\mathbf{99.8}$}}    \\
    Toothbrush   & $6.7$/$33.5$/$84.6$                                          & $4.5$/$56.1$/$75.5$                                   & {\color{blue}{$\mathbf{78.1}$}}/$95.6$/{\color{blue}{$\mathbf{99.6}$}}                        & $74.2$/{\color{blue}{$\mathbf{97.1}$}}/{\color{blue}{$\mathbf{99.6}$}}                          & {\color{red}{$\mathbf{84.8}$}}/{\color{red}{$\mathbf{98.0}$}}/{\color{red}{$\mathbf{99.7}$}}    \\
    Transistor   & $6.4$/$39.1$/$56.0$                                          & $11.0$/$49.0$/$79.1$                                  & {\color{blue}{$\mathbf{85.6}$}}/$94.8$/{\color{blue}{$\mathbf{98.4}$}}                        & $85.5$/{\color{blue}{$\mathbf{97.8}$}}/{\color{red}{$\mathbf{98.6}$}}                           & {\color{red}{$\mathbf{88.1}$}}/{\color{red}{$\mathbf{98.0}$}}/{\color{blue}{$\mathbf{98.4}$}}   \\
    Zipper       & $19.6$/$81.3$/$93.7$                                         & $42.9$/$91.0$/$96.9$                                  & $77.6$/$95.5$/$98.8$                                                                          & {\color{blue}{$\mathbf{91.0}$}}/{\color{red}{$\mathbf{99.2}$}}/{\color{blue}{$\mathbf{99.7}$}}  & {\color{red}{$\mathbf{91.6}$}}/{\color{blue}{$\mathbf{98.9}$}}/{\color{red}{$\mathbf{99.8}$}}   \\\hline
    Average      & $20.9$/$66.3$/$82.1$                                         & $22.5$/$70.1$/$82.6$                                  & $80.3$/$95.5$/$99.0$                                                                          & {\color{blue}{$\mathbf{84.3}$}}/{\color{red}{$\mathbf{98.0}$}}/{\color{blue}{$\mathbf{99.4}$}}  & {\color{red}{$\mathbf{86.8}$}}/{\color{blue}{$\mathbf{97.9}$}}/{\color{red}{$\mathbf{99.5}$}}   \\\hline
    TotalAverage & $24.4$/$71.4$/$85.3$                                         & $26.0$/$73.3$/$85.3$                                  & $78.6$/$96.1$/$99.0$                                                                          & {\color{blue}{$\mathbf{84.4}$}}/{\color{blue}{$\mathbf{98.1}$}}/{\color{blue}{$\mathbf{99.5}$}} & {\color{red}{$\mathbf{86.0}$}}/{\color{red}{$\mathbf{98.3}$}}/{\color{red}{$\mathbf{99.6}$}}    \\
    \bottomrule
  \end{tabular}
\end{table*}

\begin{table}[htbp]
  \centering
  \caption{
    The AP, PRO and Pixel-AUC scores on BTAD  \cite{mishra2021vt} in the supervised setting.
  }
  \label{table:btad_supervised_result}
  \resizebox{\linewidth}{!}{
    \begin{tabular}{lcccc}
      \toprule
      Category & \makecell{BGAD \cite{yao2022explicit} \NL (CVPR2023)}                & \makecell{PRN \cite{zhang2022prototypical} \NL (CVPR2023)} & \makecell{SemiREST \cite{li2023efficient} \NL (arXiv2023)}                                      & Ours                                                                                         \\\hline
      $01$     & $64.0$/$868.4$                                                       & $38.8$/$81.4$/$96.6$                                       & {\color{blue}{$\mathbf{81.3}$}}/{\color{blue}{$\mathbf{93.1}$}}/{\color{blue}{$\mathbf{99.0}$}} & {\color{red}{$\mathbf{82.2}$}}/{\color{red}{$\mathbf{94.0}$}}/{\color{red}{$\mathbf{99.1}$}} \\
      $02$     & $83.4$/{\color{blue}{$\mathbf{66.5}$}}/$97.9$                        & $65.7$/$54.4$/$95.1$                                       & {\color{blue}{$\mathbf{84.7}$}}/{\color{red}{$\mathbf{81.4}$}}/{\color{blue}{$\mathbf{98.1}$}}  & {\color{red}{$\mathbf{85.2}$}}/{\color{red}{$\mathbf{81.4}$}}/{\color{red}{$\mathbf{98.3}$}} \\
      $03$     & $77.4$/{\color{red}{$\mathbf{99.5}$}}/{\color{red}{$\mathbf{99.9}$}} & $57.4$/$98.3$/{\color{blue}{$\mathbf{99.6}$}}              & {\color{blue}{$\mathbf{79.9}$}}/{\color{blue}{$\mathbf{99.4}$}}/{\color{red}{$\mathbf{99.9}$}}  & {\color{red}{$\mathbf{84.6}$}}/$98.9$/{\color{red}{$\mathbf{99.9}$}}                         \\\hline
      Average  & $74.9$/$84.2$/$98.7$                                                 & $54.0$/$78.0$/$97.1$                                       & {\color{blue}{$\mathbf{82.0}$}}/{\color{blue}{$\mathbf{91.3}$}}/{\color{blue}{$\mathbf{99.0}$}} & {\color{red}{$\mathbf{84.0}$}}/{\color{red}{$\mathbf{91.4}$}}/{\color{red}{$\mathbf{99.1}$}} \\
      \bottomrule
    \end{tabular}
  }
  \vspace{-1.0em}
\end{table}

\subsection{Ablation Study}

To understand the success of CPR, we conducted a detailed ablation study to analyze the
major contributing components of the proposed method. These components are listed as
follows:
\begin{itemize}
  \item
        \textbf{Global Retrieval} process. It is the most basic component of the proposed method.
        If it is not conducted, the algorithm reduces to a standard PatchCore
        \cite{roth2022towards} algorithm.
  \item
        \textbf{FEB} module. Without this module, the final predicted anomaly score map is
        $\text{A}_{\text{mul}}$ instead of $\text{A}^{\ast}$, introduced in
        Algorithm~\ref{alg:infer}.
  \item
        \textbf{Learned LRB}. If it is not employed, one performs the local retrieval on the
        raw feature tensors generated by using DenseNet$201$, as shown in \Cref{equ:raw_fea}.
  \item
        \textbf{Multiscale} strategy. If remove this component from CPR model, only the raw
        feature tensor from ``denseblock-$1$'' of DenseNet$201$ model is used. See
        Algorithm~\ref{alg:infer} for details.
  \item
        \textbf{Remote pair weighting} strategy which is introduced in
        \Cref{subsubsec:lrb_train}. The CPR without this component treats all the sampled
        negative pairs equally.
\end{itemize}
The ablation results are summarized in \Cref{table:ablation}. One can see an obvious
increasing trend as the modules are added to the CPR model one by one. The accumulated
performance gains contributed by the modules are $1.8\%$ (Image-AUC), $1.4\%$ (Pixel-AUC),
$3.4\%$ (PRO) and $15.0\%$ (AP) respectively.

\begin{table}[htbp]  
  \centering
  \caption{
    Ablation studies on our main designs: global retrieval, foreground filter, trainable
    LRB, local region retrieval, multiscale and features weight. “I”, “P”, “R”, and “O”
    respectively refer to the five metrics of Image-AUC, Pixel-AUC, PRO, and AP.
  }
  \label{table:ablation}
  \resizebox{\columnwidth}{!}{
    \begin{tabular}{ccccccccc}
      \toprule
      \multicolumn{5}{c}{Module} & \multicolumn{4}{c}{Performance}                                                                                                                                       \\
      \cmidrule(lr){1-5} \cmidrule(lr){6-9}
      \makecell{GRB}             & \makecell{FEB}                  & \makecell{LRB} & \makecell{Mul. \NL Scale} & \makecell{Pair \NL Weight} & I $\uparrow$ & P $\uparrow$ & O $\uparrow$ & A $\uparrow$ \\
      \midrule
                                 &                                 &                &                           &                            & $97.9$       & $97.8$       & $94.4$       & $67.7$       \\
      $\checkmark$               &                                 &                &                           &                            & $94.8$       & $98.6$       & $94.9$       & $69.1$       \\
      $\checkmark$               & $\checkmark$                    &                &                           &                            & $94.9$       & $98.3$       & $93.0$       & $72.0$       \\
      $\checkmark$               &                                 & $\checkmark$   &                           &                            & $98.6$       & $98.5$       & $95.3$       & $76.1$       \\
      $\checkmark$               & $\checkmark$                    & $\checkmark$   &                           &                            & $99.0$       & $98.9$       & $97.2$       & $80.6$       \\  
      $\checkmark$               & $\checkmark$                    & $\checkmark$   & $\checkmark$              &                            & $99.7$       & $99.2$       & $97.6$       & $82.3$       \\
      $\checkmark$               & $\checkmark$                    & $\checkmark$   & $\checkmark$              & $\checkmark$               & $99.7$       & $99.2$       & $97.8$       & $82.7$       \\
      \bottomrule
    \end{tabular}
  }
\end{table}

\subsection{Algorithm Speed}

In real-world AD applications, the inspection time budget is usually limited and thus the
running speed is a crucial property of an AD algorithm. In this part of the paper, we test
the time efficiency of the proposed method as well as its two accelerated variations,
namely the ``CPR-Fast'' algorithm and the ``CPR-Faster'' algorithm, respectively. In
specific, CPR-Fast keeps employing DenseNet$201$ as the backbone while the dimension of
LRB feature is reduced to $64$ from the original $384$; CPR-Faster utilizes EfficientNet
\cite{tan2020efficientnet} as its backbone and further reduces the dimension of the LRB feature to $16$.
The running speeds of $11$ SOTA algorithms are also tested and to make a fair comparison,
all the tests are conducted on one machine and mostly based on the official code
published by the authors. In particular, we run each AD algorithm $2000$ times on the
image with the fixed size $320 \times 320$ and only the last $1000$ times are used to
estimate the average speed. To further exploit the speed potential of the proposed method, we
also adopt the TensorRT SDK \cite{migacz20178} to reimplement the CPR which is originally coded based
on Pytorch \cite{paszke2019pytorch} and the running speeds of the two frameworks are both measured.

The speeds of different algorithms are presented in \Cref{table:speed}, along with the
corresponding AD performances on the MVTec AD dataset. As we can see, CPR achieves the
highest Pixel-AUC, PRO and AP scores while maintaining a high speed (over $110$ FPS). The
CPR-Fast algorithm reaches the inference speed over $360$ FPS at a trivial cost of accuracy
drop.  Most remarkably, the CPR-Faster method achieves a speed of $1016$ FPS and still
outperform other compared SOTA methods except SemiREST \cite{li2023efficient} which requires more than $120$ ms to process an image. In other words, the CPR-Faster algorithm can
realize highly accurate anomaly detection/localization on an image using less than $1$
millisecond.

\begin{table}[htbp]
  \centering  
  \caption{
    Speed Comparison between the proposed method and current SOTA algorithms on MVTec
    AD.  “I”, “P”, “R” and “O” respectively refer to the Image-AUC, Pixel-AUC, PRO and AP. The number with symbol $^*$ stands for the result obtained by a re-implementation of the algorithm. Note that for CPR and its variations, we show the results with both the PyTorch \cite{paszke2019pytorch} implementation and the TensorRT \cite{migacz20178} version.}
  \label{table:speed}
  \resizebox{\columnwidth}{!}{
    \begin{tabular}{ c | c | c | c | c | c  }
      \hline
      Methods                                     & I $\uparrow$                    & P $\uparrow$                    & O $\uparrow$                    & A $\uparrow$                    & \makecell{FPS $\uparrow$ \NL (PyTorch/TensorRT)}             \\
      \hline
      PatchCore-WRes \cite{roth2022towards}       & $97.8$                          & $97.8$                          & $94.3$                          & $65.9$                          & $13$/$\sim$                                                  \\
      DREAM \cite{zavrtanik2021draem}             & $98.0$                          & $97.3$                          & $91.3$                          & $68.4$                          & $50$/$\sim$                                                  \\
      RD \cite{deng2022anomaly}                   & $98.5$                          & $97.8$                          & $93.9$                          & $55.4$                          & $68$/$\sim$                                                  \\
      NFAD \cite{yao2022explicit}                 & $97.7$                          & $98.2$                          & $95.2$                          & $62.5$                          & $94$/$\sim$                                                  \\
      DMAD \cite{liu2023diversity}                & $99.5$                          & $97.9$                          & $93.3$                          & $59.8$                          & $34$/$\sim$                                                  \\
      CDO \cite{cao2023collaborative}             & $96.8$                          & $98.1$                          & $94.1$                          & $57.6$                          & 9/$\sim$                                                     \\
      SimpleNet \cite{Liu_2023_CVPR}              & {\color{blue}{$\mathbf{99.6}$}} & $97.7$                          & $91.1$                          & $58.1$                          & $64$/$\sim$                                                  \\
      DeSTSeg \cite{zhang2023destseg}             & $98.6$                          & $97.9$                          & $94.4$                          & $75.8$                          & $122$/$\sim$                                                 \\
      EfficientAD-S \cite{batzner2023efficientad} & $98.8$                          & $96.8$                          & $96.5$                          & $65.9$                          & {\color{blue}{$\mathbf{394^*}$}}/$\sim$                      \\
      EfficientAD-M \cite{batzner2023efficientad} & $99.1$                          & $96.9$                          & $96.6$                          & $63.8$                          & $182^*$/$\sim$                                               \\
      SemiREST \cite{li2023efficient}             & $\sim$                          & {\color{red}{$\mathbf{99.2}$}}  & $97.5$                          & $81.2$                          & $8$/$\sim$                                                   \\
      \hline
      CPR                                         & {\color{red}{$\mathbf{99.7}$}}  & {\color{red}{$\mathbf{99.2}$}}  & {\color{red}{$\mathbf{97.8}$}}  & {\color{red}{$\mathbf{82.7}$}}  & $113$/$130$                                                  \\
      CPR-fast                                    & {\color{red}{$\mathbf{99.7}$}}  & {\color{red}{$\mathbf{99.2}$}}  & {\color{blue}{$\mathbf{97.7}$}} & {\color{blue}{$\mathbf{82.3}$}} & $245$/{\color{blue}{$\mathbf{362}$}}                         \\
      CPR-faster                                  & $99.4$                          & {\color{blue}{$\mathbf{99.0}$}} & $97.3$                          & $80.6$                          & {\color{red}{$\mathbf{478}$}}/{\color{red}{$\mathbf{1016}$}} \\
      \hline
    \end{tabular}
  }
\end{table}

\begin{figure}[htbp]
  \centering
  \includegraphics[width=3.5in]{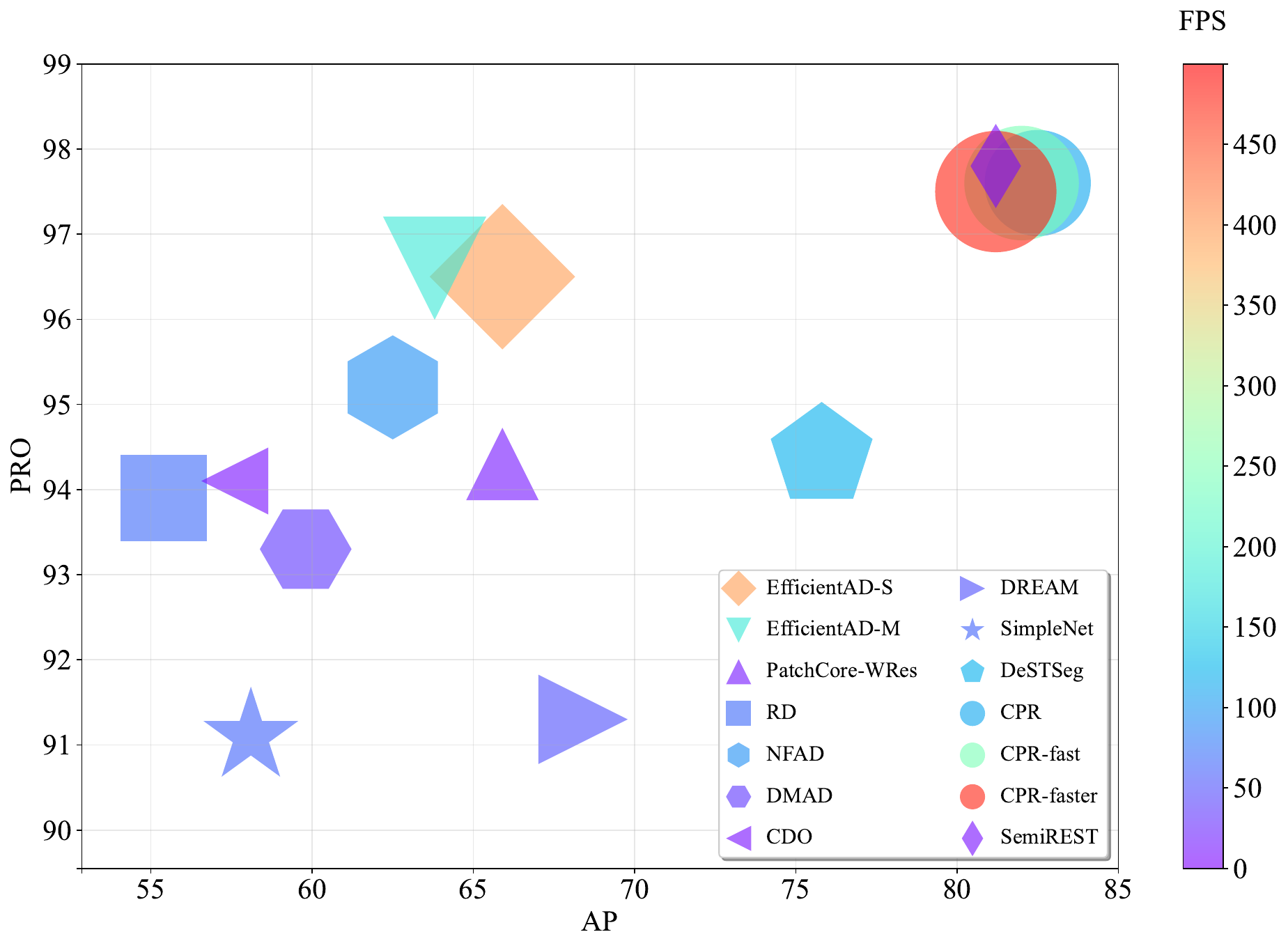}
  \caption{Inference speed (FPS) versus AP and PRO on MVTec AD benchmark. The algorithm speeds are indicated by the colors (the redder the faster) and sizes (the larger the faster) of the markers. }
  \label{fig:speed}
  \vspace{-2.0em}
\end{figure}

To better illustrate the advantage of the proposed method, \Cref{fig:speed} depicts the AD
involved AD methods as differently shaped markers with the marker locations indicating the
accuracies and the marker colors indicating the algorithm speeds.

\subsection{Qualitative Comparisons}
\label{subsec:qualitative}

In this section, qualitative AD results are shown for offering more intuitive
understanding of the proposed method.

\begin{figure*}[htbp]
  \centering
  \includegraphics[width=6in]{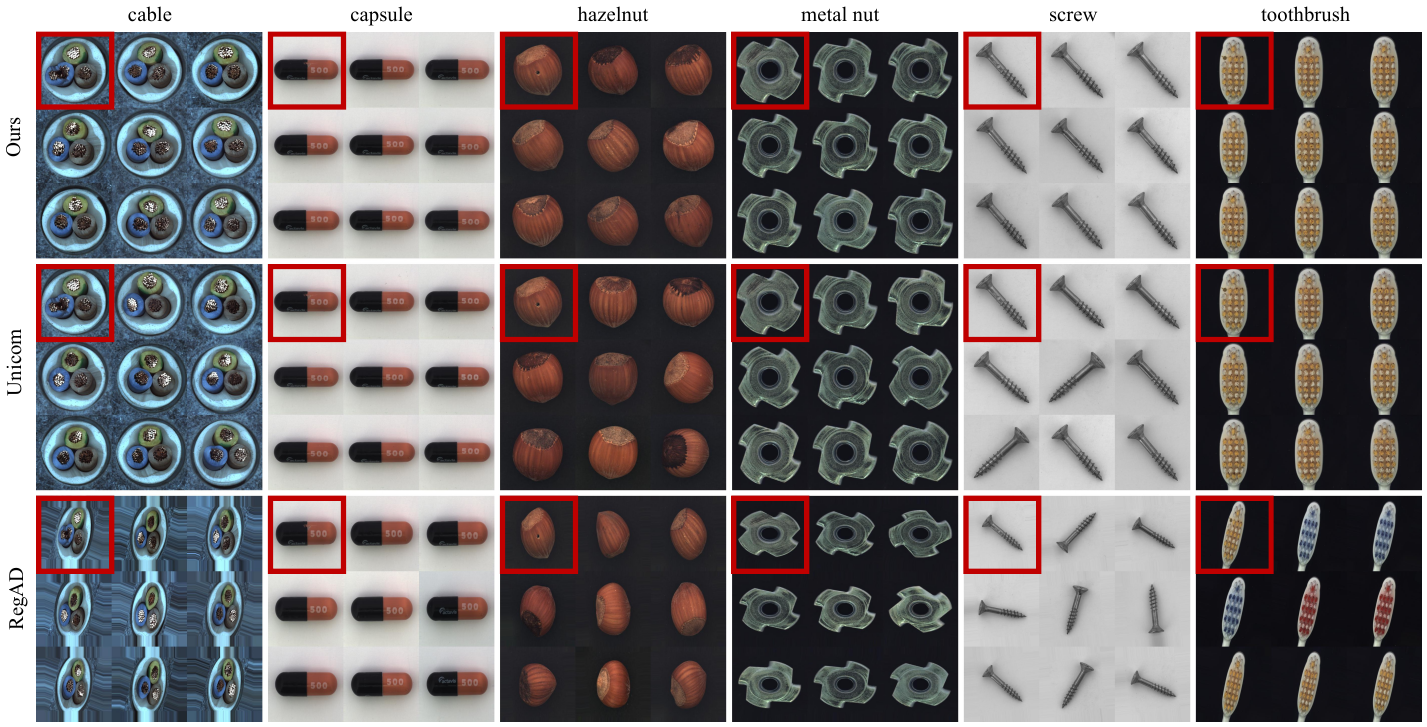}
  \caption{
    Global retrieval results on MVTec AD. The proposed GRB (top row) achieves more robust
    retrieval results for various categories of defective objects, comparing with the Unicom
    algorithm \cite{an2023unicom} (middle row) and the RegAD method \cite{huang2022registration} (bottom row).
    The images inside the red rectangles are the test images while the others are the retrieved $8$-NN images.
    Note that for RegAD, the $8$ training image are randomly selected and all the $9$ images are supposed to be aligned to similar poses according to the original paper \cite{huang2022registration}.
  }
  \label{fig:retrieval}
\end{figure*}

As a key step of CPR, the global retrieval procedure
select the ``similarly posed'' reference images for the query. \Cref{fig:retrieval}
illustrate the retrieved $8$ nearest neighbors of a typical test image (inside the red rectangle) for each category.
All the object categories in the MVTec AD dataset \cite{Bergmann_2019_CVPR} are involved in this comparison except those sub-categories with well-aligned images.
In \Cref{fig:retrieval}, our CPR method (top row) are compared with the SOTA image retrieval algorithm Unicom \cite{an2023unicom} (middle row) and RegAD \cite{huang2022registration} (bottow row) which is a SOTA AD approach employing STN layers for image alignment.
Note that RegAD is not retreival-based and performs an implicit image alignment for each test image. As a result, we show the test image (inside the red rectangle) and $8$ randomly selected training images, all aligned using the affine transform parameters predicted by the STN layers of RegAD \cite{huang2022registration}.
One can see that global retrieval process of CPR usually leads to a set of
samples which seem ``aligned'' to the test image. In practice, this merit significantly benefits the proposed CPR algorithm for both training and inference. In contrast, the general retrieval algorithm Unicom \cite{Bergmann_2019_CVPR} performs worse than CPR and the alignment-based RegAD \cite{huang2022registration} fails to align all the images to similar poses.

\begin{figure*}[htbp]
  \centering
  \includegraphics[width=5in]{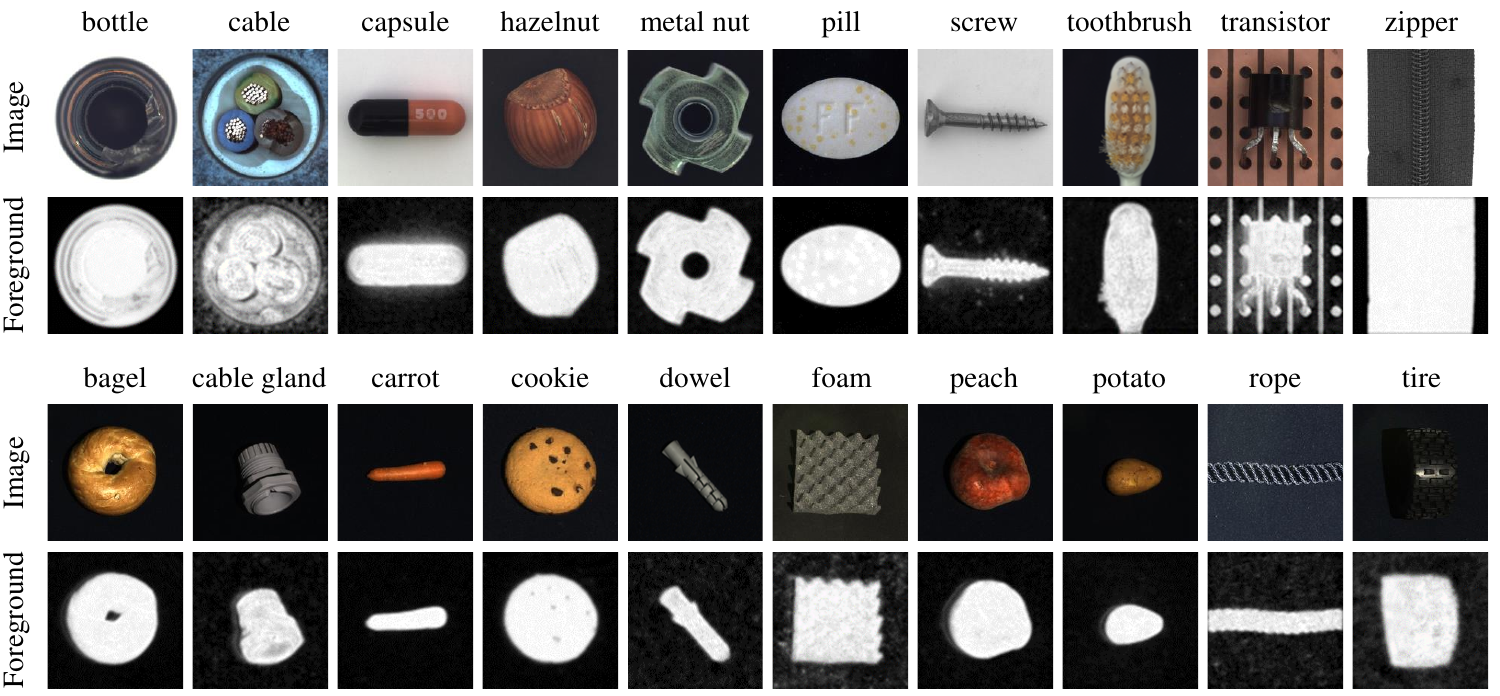}
  \caption{
    The demonstration of the proposed FEB module. All the object categories in MVTec AD (top row) and MVTec-3D AD (bottom row) are illustrated in this figure. Note that instead of the binary map, CPR utilizes the foreground confidence to reduce the influence of the ``false alarms'' in the background region.
  }
  \label{fig:foreground_vis}
\end{figure*}

Secondly, in \Cref{fig:foreground_vis} we demonstrate the foreground estimation of the proposed FEB module. As it can be seen, the FEB can accurately predict the pixel labels (foreground \emph{vs.} background) even though no manual annotation is available in the training stage.

\begin{figure*}[htbp]
  \centering
  \includegraphics[width=7in]{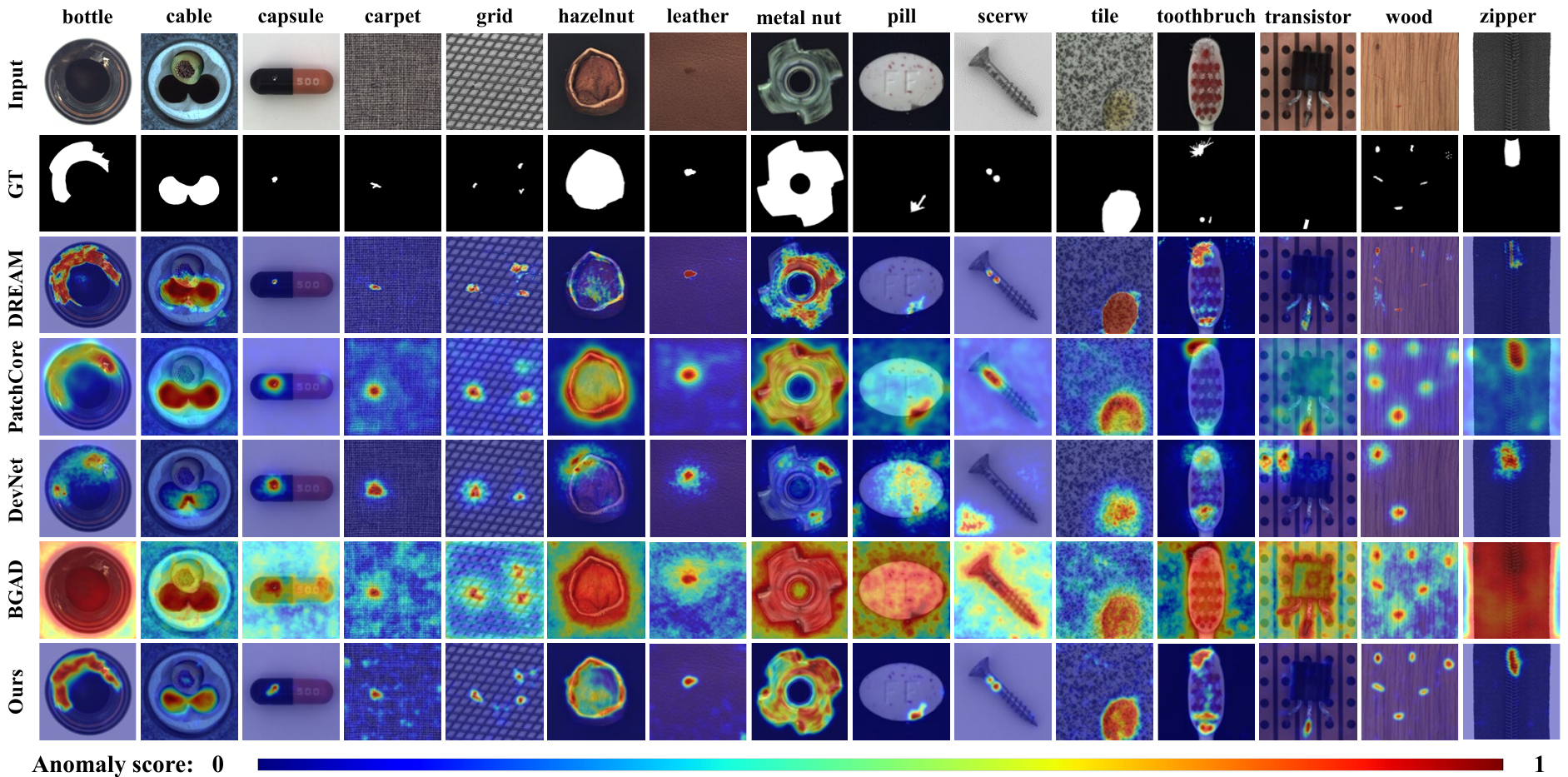}
  \caption{
    Visualization comparisons examples of anomaly localization on MVTec AD. Our method
    accurately localizes anomalous regions and shows more similar predictions to the ground
    truth.
  }
  \label{fig:vis}
\end{figure*}

Finally, the anomaly score maps of our method and $4$ representative AD algorithms are
shown in \Cref{fig:vis}. As we can see from the image, CPR predicts more accurate
score maps in most scenarios.

\section{Conclusion \& future work}
\label{sec:conclusion}
In this paper, we rethink the inherent ``matching'' nature of anomaly detection and consequently
propose to perform AD in a cascade manner. The former layer of the cascade filters out
most improper reference images while the test patch is matched with the reference set in
the latter layer. This coarse-to-fine retrieval strategy is proved to be remarkably useful
and yields new SOTA AD accuracy as well as a running speed over $1000$ FPS with a
moderate simplification. Besides the good performances, this work also sheds light on
the understanding of all the matching-based AD algorithms. According to the empirical
study of this work, the high accuracy usually stems from two crucial factors, namely the
reference set selection and the metric feature learning. In the future, 
algorithms
can be 
improved
by finding more sophisticated alternatives to those two key components.
In addition, the running speed of the retrieval-based AD process could be further
increased by introducing the hashing 
techniques which have wide applications in
fast
retrieval.

\bibliography{strings,ref}
\bibliographystyle{IEEEtran}

\end{document}